\documentclass{sig-alternate-10pt-compact}[10pt]
\usepackage{epsfig,endnotes}
\usepackage[usenames,dvipsnames]{color}
\usepackage{xspace}
\usepackage{times}
\usepackage{amsmath}
\usepackage{url}
\usepackage{gensymb}
\usepackage{pifont}
\usepackage[linesnumbered,ruled,vlined]{algorithm2e}
\usepackage{microtype}
\usepackage{graphicx}
\usepackage{subcaption}
\usepackage{multirow}
\usepackage{textcomp}

\begin{document}

\newcommand{\BULLET}{\vspace{+.00in} \noindent $\bullet$ \hspace{+.00in}}
\renewcommand{\paragraph}[1]{\noindent{\bf{#1}}}
\renewcommand{\ttdefault}{cmtt}
\renewcommand{\sfdefault}{cmss}

\newcommand{\hd}[1]{\small{\textbf{\texttt{#1}}}\normalsize}
\newcommand{\hds}[1]{\small{\textbf{\texttt{#1}}}}
\newcommand{\hdf}[1]{\small{\textbf{\texttt{#1}}}\footnotesize}

\newcommand{\etc}{\textit{etc.}\xspace}
\newcommand{\ie}{\textit{i.e.,}\xspace}
\newcommand{\eg}{\textit{e.g.,}\xspace}
\newcommand{\etal}{\textit{et al.}\xspace}
\newcommand{\wrt}{\textit{w.r.t.}\xspace}
\newcommand{\aka}{\textit{a.k.a.}\xspace}

\newcommand{\feng}[1]{{\color{red}[\textsf{#1}]}}

\newcommand{\mycomment}[1]{{\color{red}[\textsf{#1}]}}

\newcommand{\DEG}{\degree\xspace}

\newcommand{\mysection}[1]{\vspace{-.05in}\section{#1}\vspace{-.02in}}
\newcommand{\mysubsection}[1]{\vspace{-.05in}\subsection{#1}\vspace{-.02in}}
\newcommand{\mysubsubsection}[1]{\vspace{-.05in}\subsubsection{#1}\vspace{-.02in}}

\setlength{\emergencystretch}{1em}

\date{}

\thispagestyle{empty}

\newcommand{\name}{$\sf\small{DeepMix}$\xspace}	
\newcommand{\boldname}{\textbf{DeepMix}\xspace}
\newcommand{\bigname}{DeepMix\xspace}

\newcommand{\bo}[1]{{\color{red}#1}}

\title{\bigname: Mobility-aware, Lightweight, and Hybrid 3D Object Detection for Headsets}

\author{%
Yongjie Guan\\
  \affaddr{New Jersey Institute of Technology}\\
  \affaddr{Newark, NJ, USA}\\
  \email{yg274@njit.edu}
\and
Xueyu Hou\\
  \affaddr{New Jersey Institute of Technology}\\
  \affaddr{Newark, NJ, USA}\\
  \email{xh29@njit.edu}
\and
Nan Wu\\
  \affaddr{George Mason University}\\
  \affaddr{Fairfax, VA, USA}\\
  \email{nwu5@gmu.edu}
\and
Bo Han\\
  \affaddr{George Mason University}\\
  \affaddr{Fairfax, VA, USA}\\
  \email{bohan@gmu.edu}
\and
Tao Han\\
  \affaddr{New Jersey Institute of Technology}\\
  \affaddr{Newark, NJ, USA}\\
  \email{tao.han@njit.edu}
}

\maketitle

\begin{abstract}

Mobile headsets should be capable of understanding 3D physical environments to offer a truly immersive experience for augmented/mixed reality (AR/MR).
However, their small form-factor and limited computation resources make it extremely challenging to execute in real-time 3D vision algorithms, which are known to be more compute-intensive than their 2D counterparts.
In this paper, we propose \name, a {\em mobility-aware, lightweight, and hybrid} 3D object detection framework for improving the user experience of AR/MR on mobile headsets.
Motivated by our analysis and evaluation of state-of-the-art 3D object detection models, \name intelligently combines {\em edge-assisted 2D object detection} 
and novel, {\em on-device 3D bounding box estimations} that  leverage depth data captured by headsets.
This leads to low end-to-end latency and significantly boosts detection accuracy in mobile scenarios.
A unique feature of \name is that it fully {\em exploits the mobility of headsets to fine-tune detection results and boost detection accuracy}. 
To the best of our knowledge, \name is the first 3D object detection that achieves 30 FPS (\ie an end-to-end latency much lower than the 100 ms stringent requirement of interactive AR/MR).
We implement a prototype of \name on Microsoft HoloLens and evaluate its performance via both extensive controlled experiments and a user study with 30+ participants.
\name not only improves detection accuracy by 9.1--37.3\% but also reduces end-to-end latency by 2.68--9.15$\times$, compared to the baseline that uses existing 3D object detection models.  

\end{abstract}

\mysection{Introduction}

Mobile headsets such as Microsoft HoloLens~\cite{HoloLens2} and Magic Leap One~\cite{magicleap} bring numerous opportunities to enable truly immersive augmented/mixed reality (AR/MR).
To offer the best quality of experience (QoE), real-time, interactive AR/MR  should be able to perceive and understand the surrounding environment in 3D for seamlessly blending virtual and physical objects~\cite{choi2019lpgl}.
%
%
%
%
With recent advances in 3D data capturing devices such as LiDAR and depth cameras, the computer vision (CV) community has developed several efficient 3D object detection algorithms~\cite{DBLP:conf/cvpr/DingHYWSLL20,lang2019pointpillars,DBLP:conf/cvpr/QiSMG17,shi2020pvrcnn,DBLP:conf/iccv/TangL19,DBLP:conf/cvpr/XieLWWZXW20,DBLP:conf/cvpr/ZhouT18} by leveraging deep neural networks (DNNs).
Due to the huge amount of data to process, 3D object detection is more computation-intensive than its 2D counterpart~\cite{feng2020mesorasi}.
Moreover, the performance of 3D vision algorithms heavily depends on the quality of input data (\eg  point cloud density or depth image resolution)~\cite{zhang2020mobile}.
Thus, existing AR/MR systems~\cite{apicharttrisorn2019frugal,Chen:2018:MEM,Liu:2019:EAR,zhang2018jaguar} mainly focus on 2D object detection. 

Even for the 2D case, it is well-known that the high latency caused by DNN inference negatively impacts the quality of user experience~\cite{apicharttrisorn2019frugal}.
A widely-used acceleration technique is to offload the compute-heavy tasks to cloud/edge servers~\cite{Liu:2019:EAR,zhang2018jaguar}, which is also a promising solution to speed up 3D object detection.
However, we find that even with the help of a powerful GPU, the inference time of 3D object detection ranges from 72 to 283 ms (\S\ref{sec:challenge}).
By considering the network latency for offloading and local processing time on headsets, the end-to-end latency of AR/MR systems that integrate existing 3D object detection models would be higher than 100 ms, the threshold required by interactive AR/MR~\cite{Chen:2018:MEM,Liu:2019:EAR}, hindering providing an immersive experience to users.

\begin{table}[]
\centering
\small
\begin{tabular}{c|cccc}
Methods& Input & Lat. & Accuracy & Mobility\\
\hline 
MLCVNet~\cite{DBLP:conf/cvpr/XieLWWZXW20} & Point Cloud & High & Medium & $-$\\
Trans3D~\cite{DBLP:conf/iccv/TangL19} & RGB-D  & Med. High & Low & $-$\\
D$^4$LCN~\cite{DBLP:conf/cvpr/DingHYWSLL20} & RGB  & Med. Low & Medium & $-$\\
\hline
DeepMix & Hybrid & Low & High & $+$\\
\end{tabular}
\vspace{-0.1in}
\caption{\small 
Comparison of \name and existing DNN-based 3D object detection. By exploiting headset mobility ($+$), \name achieves low end-to-end latency and high detection accuracy. }
\label{tab:art}
\vspace{-0.31in}
\end{table}

In this paper, we propose \name, a mobility-aware, lightweight, and accurate 3D object detection framework that can offer 30 frames per second (FPS) processing rate on Microsoft HoloLens 2, a commodity mobile headset.
Our key insight is that instead of utilizing DNN-based 3D object detection models to learn object class and infer bounding box, we can {\em decouple} the whole process and measure/estimate the 3D bounding box of an object by processing depth data on headsets.
The key challenge of designing \name is again the huge amount of 3D data to handle, given the limited computation resources on the headset.
Also, while it is feasible, although not trivial, to measure the size and the 6DoF (six degrees of freedom) pose of an object (\ie its position and orientation), we still need to label the object of interest.

To address the above challenges, we design a {\em hybrid mechanism} that combines the mature DNN-based 2D object detection, which is fast by offloading it to the edge, and our lightweight and intelligent on-device depth data processing.
More specifically, \name offloads only 2D RGB images to the edge for object detection (\ie getting the label) and benefits from the returned 2D bounding box to drastically reduce the amount of to-be-processed 3D data.
By doing this, \name achieves accurate 3D object detection at line-rate (\ie 30 FPS).
A unique feature of \name is that it can fully exploit the movement of users to further fine-tune the measured bounding box and boost object-detection accuracy. 
To the best of our knowledge, \name is the first 3D object detection framework that can bring about both low latency and high accuracy.
We compare \name and existing DNN-based models in Table~\ref{tab:art}.

Our detailed study of \name consists of the following.

{\bf Performance Dissection of Existing 3D Object Detection Methods (\S\ref{sec:background}).}
To understand the feasibility of applying existing DNN-based 3D object detection to interactive AR/MR, we investigate the detection accuracy and computation latency of eight state-of-the-art algorithms.
%
%
We find that existing methods are not ready for real-time AR/MR applications due to the high computation latency.
%
%

\begin{figure}[t]
    \centering
      \includegraphics[width=0.75\linewidth]{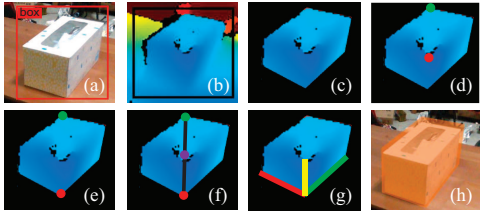}
	    \vspace{-0.1in}
	   
	   \caption{\small Workflow of \name: (a) 2D bounding box on an image, (b) Bounding box alignment on a depth frame, (c) Background removal, (d) Key points detection, (e) Key point projection, (f) Central point calculation, (g) Dimension and orientation estimation, (h) 3D bounding box visualization. }
	    \label{fig:workflow}
	    \vspace{-0.3in}
\end{figure}

{\bf Novel System Design of \name (\S\ref{sec:design}).}
%
%
As shown in Figure~\ref{fig:workflow}, \name starts by offloading only RGB images to the edge that executes 
2D object detection models for labeling objects of interest and generating their 2D bounding boxes (Figure~\ref{fig:workflow} (a)).
After aligning the bounding box on the depth image, it extracts only depth data of the target object (Figure~\ref{fig:workflow} (b)--(c)).
It then detects two key points on the 3D bounding box and projects one of them to the ground for determining the center point of the box (Figure~\ref{fig:workflow} (d)--(f)).
Finally, after inferring the dimension of the object and measuring its orientation, \name renders the 3D bounding box on the display of the headset (Figure~\ref{fig:workflow} (g)--(h)).

{\bf Effective Performance Optimization of \name (\S\ref{sec:optimization}).}
To further improve detection accuracy and end-to-end latency, we propose a few optimizations for \name.
Our key optimization is to leverage device mobility 
to refine the estimated bounding box.
By doing this, we dramatically enhance the detection accuracy of \name in dynamic environments.
This feature makes \name competitive for mobile AR/MR.
%

{\bf Implementation of \name and Performance Evaluation (\S\ref{sec:results}).}
We build a prototype implementation of \name and thoroughly
evaluate its performance via repeatable, controlled (live) experiments and a user study with more than 30 participants. 
We highlight our evaluation results as follows.

\BULLET On a high-throughput WiFi network, the end-to-end latency of \name is only 34 ms (\S\ref{sec:latency}), much lower than that of existing DNN-based models (ranging from 91 to 311 ms).

\BULLET Compared to the besting performing existing model, the accuracy improvement of \name increases from 3.5\% for the static scenarios to up to 9.1\% for the mobile scenarios.
%
%

\BULLET The experimental results from our user study demonstrate that the accuracy of \name is 12.5\%, 5.1\%, and 9.6\% higher than that of the most accurate existing model for three pre-defined mobility patterns, leading to a better QoE (\S\ref{sec:user-study}).

Overall, \name is a first-of-its-kind practical 3D object detection framework for mobile headsets.
We make the following contributions in this paper: 
(1) performance dissection of DNN-based 3D object detection models in the context of real-time, interactive AR/MR on mobile headsets; 
(2) system design of \name, a full-fledged, ready-to-deploy 3D object detection framework for commodity mobile headsets that fully exploits device mobility to boost detection accuracy; 
and (3) prototype implementation and evaluation of \name, including dataset-driven repeatable experiments, controlled live experiments, and an IRB-approved user study. 
%
%
We plan to release the implementation of \name. 


\mysection{Background \& Motivation}
\vspace{-0.1in}
\label{sec:background}

\begin{figure}[t]
\includegraphics[width=0.85\linewidth]{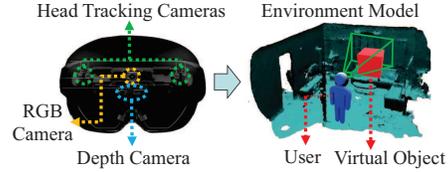}
\caption{\small Configuration of mobile headsets (\ie Microsoft HoloLens 2) and a typical application scenario.}
\label{fig:MR_operations}
\end{figure}

\subsection{Mobile Headsets for AR/MR}

Different from smartphones that can support only video see-through AR by overlaying virtual content in the physical world that is displayed via the devices' camera view, headsets allow users to see the physical world through a transparent, optical see-through display that simultaneously imposes virtual objects into the user's view of the surrounding environment using optical combiners~\cite{Grubert:2018:SCM}. 
As a result, those headsets create a truly immersive AR/MR experience, compared to smartphones and tablets, by extending our perception of the environment from 2D image frames to the 3D real world and enabling interactions between users and virtual objects.
%

Take Microsoft HoloLens as an example~\cite{Hololens:MR}. 
As illustrated in Figure~\ref{fig:MR_operations}, it has an RGB camera, a time-of-flight (ToF) sensor for depth perception, four visible light cameras for head tracking, and two infrared cameras for eye tracking. 
It is also equipped with an inertial measurement unit (IMU) with an accelerometer, gyroscope, and magnetometer. 
With these sensors, HoloLens can perceive the surrounding environment by building a 3D model and blending the digital and physical worlds based on this 3D model of the environment. 
To accurately mix virtual content with physical objects, HoloLens creates a spatial coordinate system of the physical world. 
This coordinate system uses the initial location where the HoloLens was turned on as the origin.
Moreover, to guarantee an immersive experience, AR/MR applications running on HoloLens should be capable of detecting objects in 3D space (\ie conducting 3D object detection~\cite{lang2019pointpillars,DBLP:conf/cvpr/QiSMG17,shi2020pvrcnn,DBLP:conf/cvpr/ZhouT18}), instead of leveraging 2D object detection in traditional AR systems~\cite{apicharttrisorn2019frugal,Chen:2018:MEM,Liu:2019:EAR,zhang2018jaguar}.

Mobile headsets are usually lightweight and wearable. 
As a result, their hardware resources and computation capabilities are limited. 
For instance, Microsoft HoloLens has an Intel 1GHz 32-bit processor with a customized holographic processing unit (HPU) and only 2GB of memory~\cite{Hololens:MR}. 
Such limited computation resources make it challenging to support the real-time execution of deep neural networks for 3D object detection~\cite{lang2019pointpillars,DBLP:conf/cvpr/QiSMG17,shi2020pvrcnn,DBLP:conf/cvpr/ZhouT18}. 
Furthermore, headsets' batteries can usually last only 2-3 hours, and the heat generated from the headset cannot only be dissipated via passive cooling. 
Hence, considering the power consumption and device surface temperature, mobile headsets are unsuitable for executing heavy computation tasks.

\mysubsection{A Primer of 3D Object Detection}

%
We can classify existing methods into three categories based on their input-data format. 
The first one utilizes point clouds as the input and directly draws 3D bounding boxes on them~\cite{lang2019pointpillars,qi2019deep,Qi_2020_CVPR,ren2016three,shi2020pvrcnn,DBLP:conf/cvpr/XieLWWZXW20}.
Point clouds can be directly captured by LiDAR devices or generated by processing the RGB images and their corresponding depth images.
%
%
The second category uses RGB images as the input of DNN models and learns 3D bounding boxes that will be drawn on 2D images~\cite{chen20153d,DBLP:conf/cvpr/ChenKZMFU16,DBLP:conf/cvpr/DingHYWSLL20,DBLP:conf/iccv/MaWLZOF19}.
Some of the algorithms actually generate/estimate depth maps from RGB images to train the DNN models~\cite{DBLP:conf/cvpr/DingHYWSLL20,DBLP:conf/iccv/MaWLZOF19}.
%
%
Image-based 3D object detection is an active research area because its run-time inference relies on only RGB images that are much easier to capture at a low cost, compared to 3D data such as depth maps and point clouds.
The third category benefits from 2.5D data (\ie RGB-D images) that combine 2D RGB images and depth maps~\cite{lahoud20172d,qi2017frustum,song2016deep,DBLP:conf/iccv/TangL19}.
%
%
%
%
While both \name and methods in this category use RGB images and depth data, the key difference is that \name offloads only RGB images to the edge for 2D object detection and processes depth data on the headset, whereas models with RGB-D input offload both RGB and depth images for 3D object detection.
We offer a detailed review of existing work on 3D object detection in \S\ref{sec:related}.

\subsection{Challenges of 3D Object Detection}
\label{sec:challenge}

There are several challenges when executing 3D object detection for AR/MR applications on mobile headsets.
The first one is that the performance of 3D vision algorithms heavily depends on the quality of input data (\ie the resolution of depth images or the density of point clouds).
For example, the accuracy of 3D semantic segmentation decreases when the point clouds become sparse~\cite{zhang2020slimmer}.
However, due to the limited hardware resources on mobile headsets, the resolution of their captured depth images is usually low, for instance, 360$\times$360 for Microsoft HoloLens 2 at 30+ frames per second (FPS)\footnote{While Microsoft HoloLens 2 can generate 1024$\times$1024 depth images, the frames rate is only 1-5 FPS at this high resolution, which is too low for interactive AR/MR.}.
In contrast, standalone RGB-D cameras such as Intel RealSense and Microsoft Kinect DK can capture depth images with 1024$\times$768 and 1024$\times$1024 resolutions at 30+ FPS\footnote{Intel RealSense captures rectangle depth images, whereas Microsoft HoloLens and Kinect capture square ones.}.
Thus, the density of point clouds generated from the depth images is also low (\eg around only 33.5K points when using 360$\times$240 depth images).

\begin{figure}[t]
  \includegraphics[width=\linewidth]{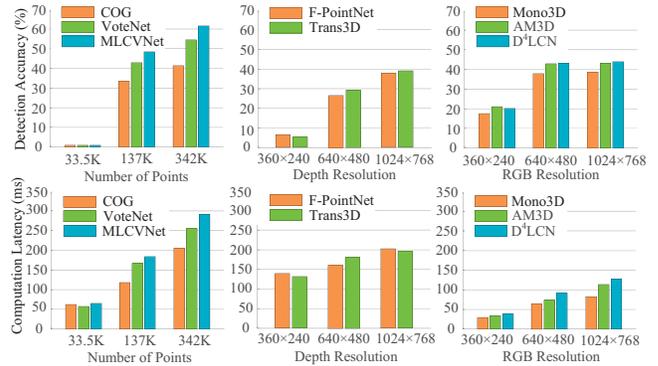}
  \vspace{-0.3in}
  \caption{\small Detection accuracy (1st row) and computation latency (2nd row) of 3D object detection methods using point cloud (left column), RGB-D (middle column), and 2D image (right column) input-data formats. The point clouds with 33.5K, 137K, and 342K points are generated from depth images with 360$\times$240, 640$\times$480, and 1024$\times$768 resolutions, respectively.}

  \label{fig:depth}
  \vspace{-0.25in}
\end{figure}

\begin{figure*}
  \centering
    \begin{minipage}{.3\linewidth}
      \centering
      \vspace{-0.15in}
      \includegraphics[width=0.80\linewidth]{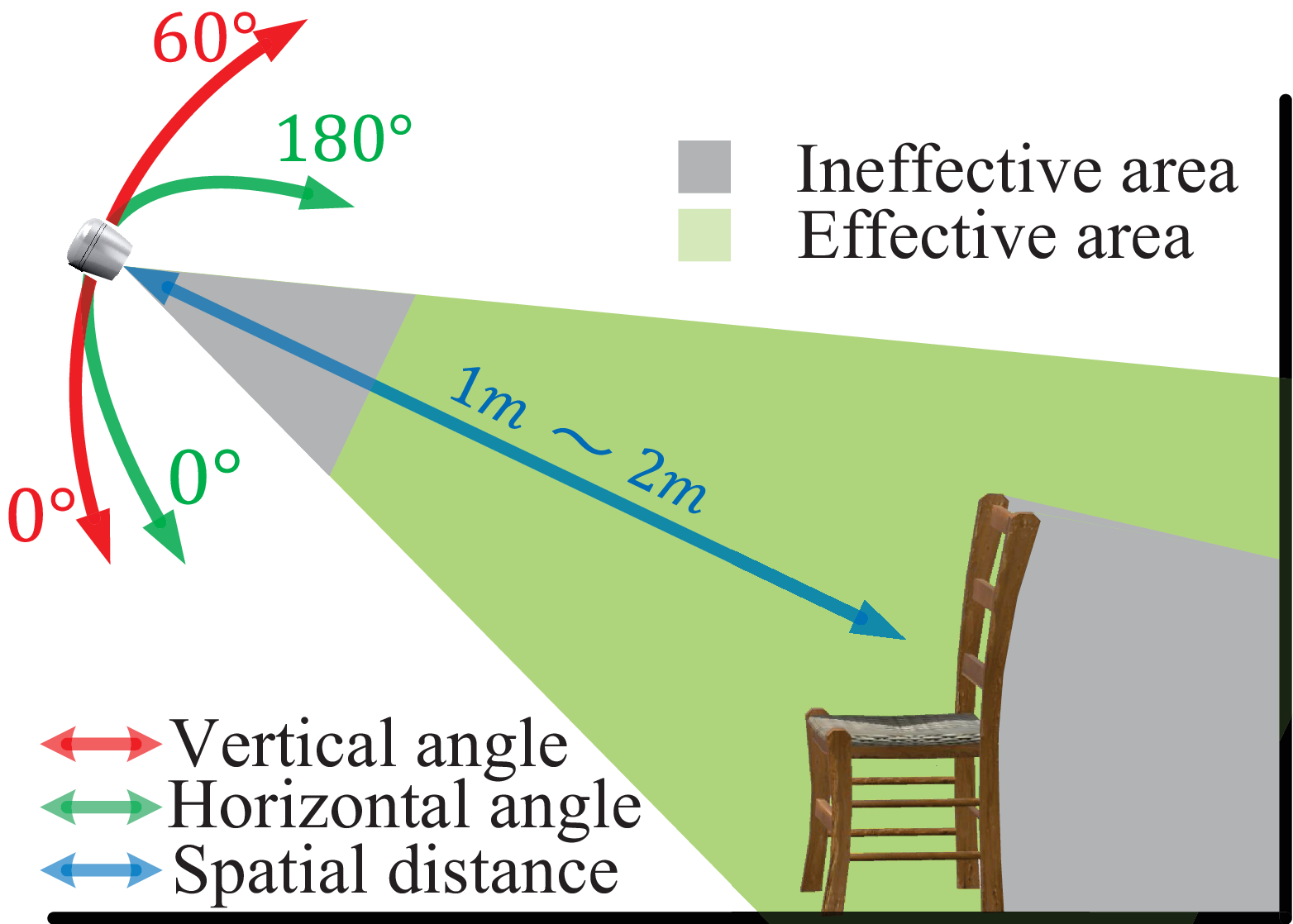}
      \vspace{-0.10in}
      \caption{\small Different settings for \\ constructing the dataset.}
      \label{fig:setting}
      \vspace{-0.24in}
    \end{minipage}%
    \begin{minipage}{.39\linewidth}
      \centering
      \includegraphics[width=0.88\linewidth]{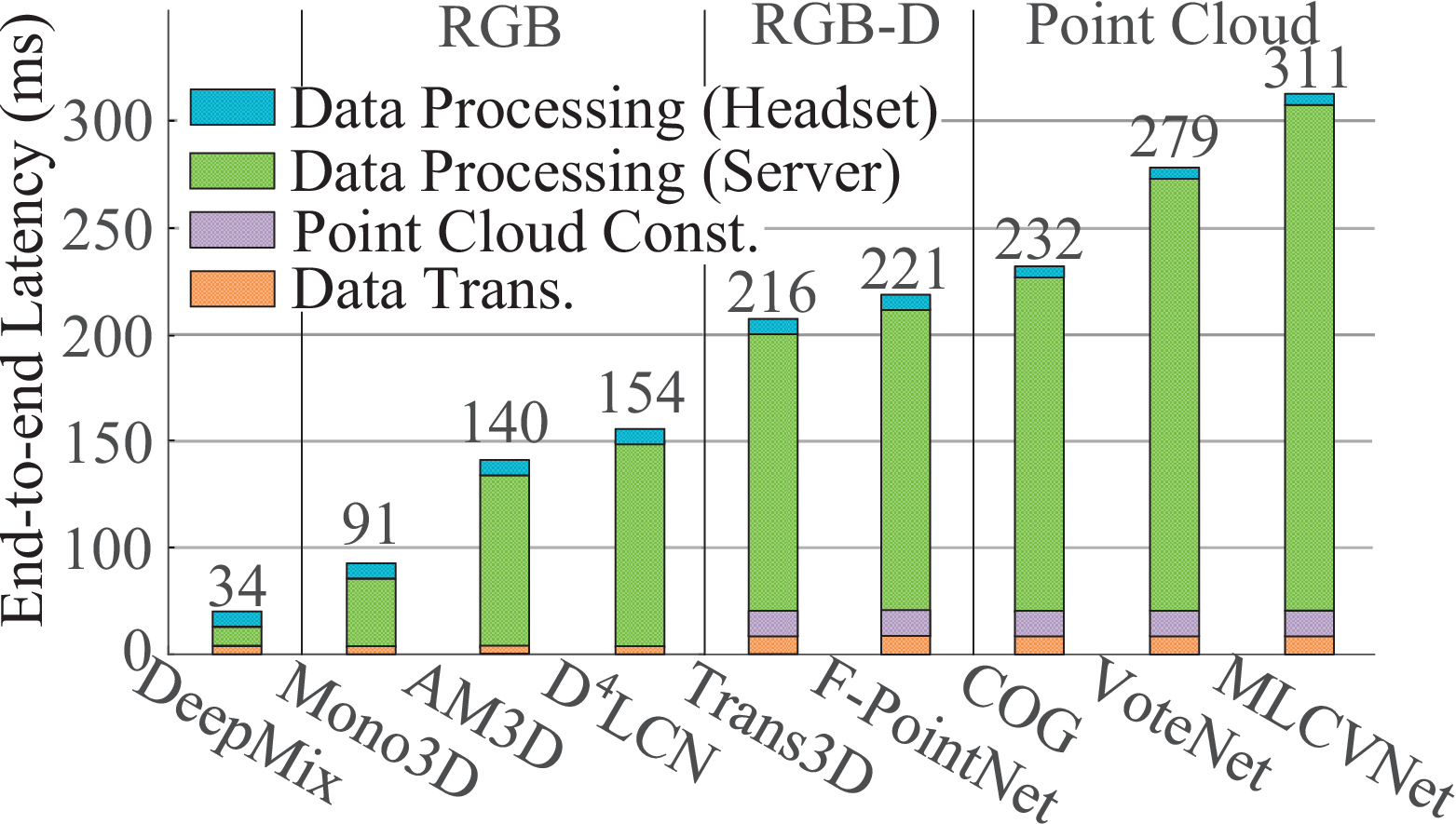}
      \vspace{-0.2in}
      \caption{\small Comparison of end-to-end latency. 
      }
      
      \label{fig:e2e}
      \vspace{-0.05in}
    \end{minipage}%
    \begin{minipage}{.3\linewidth}
      \centering
      \vspace{-0.15in}
      \includegraphics[width=0.8\linewidth]{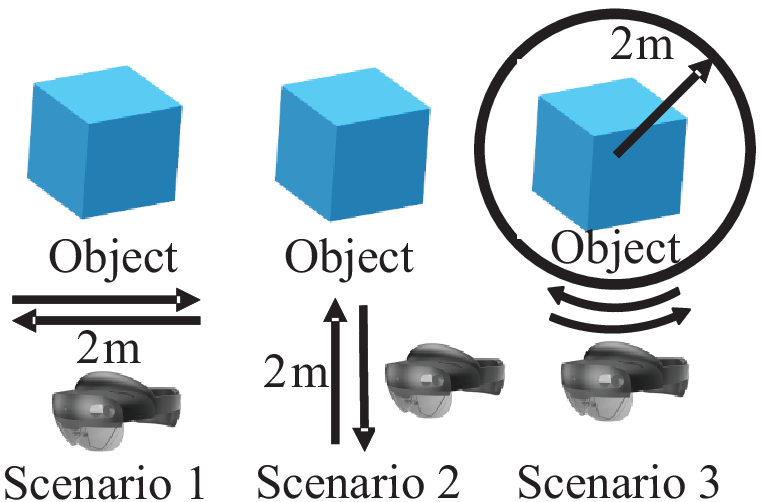}
      \caption{\small Three mobility patterns.} 
      \label{fig:patterns}
      \vspace{-0.3in}
    \end{minipage}%
\end{figure*}

In order to understand the impact of input-data quality on 3D object detection, we evaluate the accuracy of the following representative algorithms using point clouds with different densities that are generated from depth images with different resolutions.
We use 3D IoU that is defined in \S\ref{sec:results} as the evaluation metric.
We select COG~\cite{ren2016three}, VoteNet~\cite{qi2019deep}, and MLCVNet~\cite{DBLP:conf/cvpr/XieLWWZXW20} for 3D object detection with point clouds as input and F-PointNet~\cite{qi2017frustum} and Trans3D~\cite{DBLP:conf/iccv/TangL19} for models using RGB-D input.
As the baseline, we evaluate the performance of  Mono3D~\cite{DBLP:conf/cvpr/ChenKZMFU16}, AM3D~\cite{DBLP:conf/iccv/MaWLZOF19}, and D$^4$LCN~\cite{DBLP:conf/cvpr/DingHYWSLL20} that take 2D images as input.
We train the above models with the publicly available SUN RGB-D dataset~\cite{song2015sun}.
The testing RGB and depth images with different resolutions are created by the Intel RealSense camera.
As a motivating example, the captured object is a bottle under a given setting (\ie a specific viewing angle and distance, see Figure~\ref{fig:setting}).
We conduct an extensive evaluation for different objects under different settings in \S\ref{sec:results}.

The main observations from the experimental results in Figure~\ref{fig:depth} are as follows.
First, the detection accuracy is extremely low for point-cloud-based models (at most 1.6\% in the upper-left subfigure) and models using RGB-D images as input data (5.1\%-6.8\% in the upper-middle subfigure), when the resolution of depth images is 360$\times$240 (\ie the typical setup of HoloLens 2). 
Second, when the quality of input data is low, image-based models achieve the most accurate detection among the three categories, whereas point-cloud-based models are more accurate than the other two for high-quality input data.
%
%
%
Third, with high-quality input, point-cloud-based models achieve the most accurate detection, but lead to the highest computation latency.
%
%
Fourth, the computation latency of point-cloud-based and image-based models increases for high-density point clouds and high-resolution images.

Another challenge of leveraging 3D object detection for mobile AR/MR applications is the high computation overhead and the resulting high latency of data processing.
To better appreciate this issue, we compare the inference time of traditional 2D object detection models such as YOLOv4~\cite{DBLP:journals/corr/abs-2004-10934} with the aforementioned representative 3D object detection models.
The input RGB images of both 2D and 3D models have the same resolution of 1280$\times$720, to make the comparison fair.
The resolution of the input depth images is 1024$\times$768 for 3D models.
To follow the common practice of edge-based acceleration for 2D object detection/recognition in mobile AR~\cite{Liu:2019:EAR,zhang2018jaguar}, 
we conduct the experiments on a machine with an NVIDIA RTX 2080S GPU. 

\begin{figure}[t]
  \includegraphics[width=0.94\linewidth]{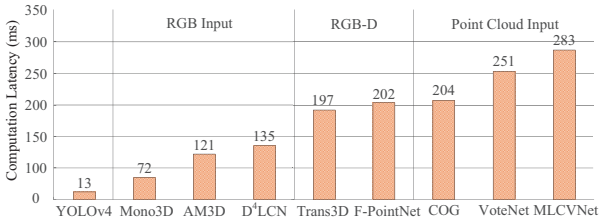}
  \vspace{-0.15in}
  \caption{\small Comparison of the computation latency of 2D vs. 3D object detection algorithms. YOLOv4 is for 2D object detection, and the rest are all for 3D object detection with different input data formats.}
  \label{fig:latency}
  \vspace{-0.3in}
\end{figure}

We have the following three observations from 
Figure~\ref{fig:latency}.
First, the computation latency for most 3D object detection algorithms is higher than 100 ms, making them unsuitable for real-time, interactive AR/MR applications~\cite{Chen:2018:MEM,Liu:2019:EAR}.
Ideally, the latency should be at most 33--34 ms to achieve 30 FPS line-rate processing.
While the latency of image-based models could be lower than 100 ms, 
as we will show in \S\ref{sec:results}, by adding the extra network latency and local computation time, the end-to-end latency would still deteriorate the quality of user experience.
Second, the computation latency of 3D object detection is much higher than its 2D counterpart.
It takes only 13 ms for YOLOv4~\cite{DBLP:journals/corr/abs-2004-10934} to detect objects on 2D images, whereas the computation latency could be as high as 283 ms for 3D models.
Third, the computation latency of 3D object detection heavily relies on the complexity of input data.
%

The above large performance gap makes edge-side optimizations, such as DNN-model acceleration and better GPU support, challenging.
Note that we assume point clouds will be created on the server to reduce network latency and computation overhead on the headsets.
Generating high-fidelity point clouds, which is required to improve detection accuracy (Figure~\ref{fig:depth}), also takes time and will further increase the latency of point-cloud-based models.

\begin{figure*}[!htp]
    \centering
    \vspace{-0.15in}
      \includegraphics[width=0.73\linewidth,height=2.25in]{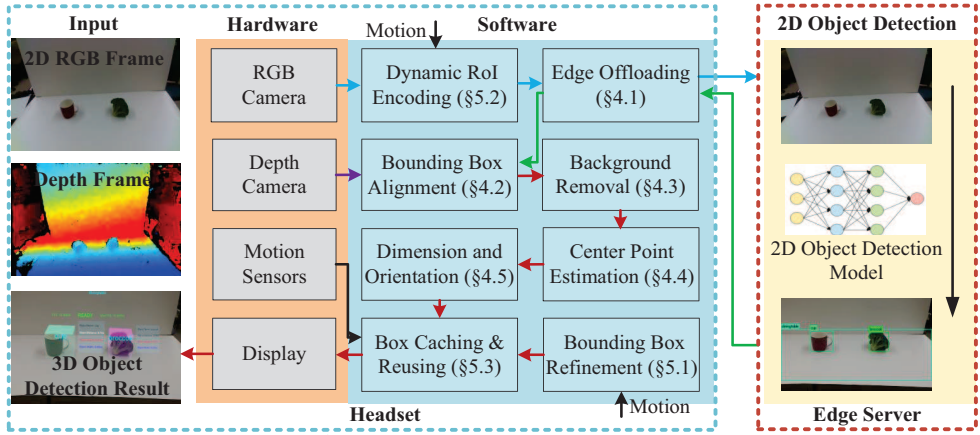}
	    \vspace{-0.15in}
	   	\caption{\small System architecture and workflow of \name.}
	    \label{fig:architecture}
	    \vspace{-0.2in}
\end{figure*}

\noindent{\bf Summary:} The state-of-the-art 3D object detection solutions are not suitable for supporting AR/MR applications on mobile headsets due to the following two reasons.

\BULLET Existing 3D object detection models achieve the most accurate result when using high-quality point clouds as input data, which {\em cannot be generated by commodity mobile headsets} due to their limited hardware resources.

\BULLET The computation latency of existing 3D object detection models, even with edge offloading, are {\em too high to guarantee a truly immersive experience} for real-time, interactive AR/MR that requires imperceptible latency ($<$100 ms).

The poor performance of existing 3D object detection models and their complex interplay among the input-data quality, detection accuracy, and computation latency motivate our design of \name, which effectively combines edge-assisted 2D object detection and on-device lightweight 3D bounding box estimation with depth data.

\mysection{Overview of \bigname}

\label{sec:sys_design}

\name is a generic 3D object detection framework that is designed for enhancing AR/MR experience on mobile headsets.
It is mobility-aware by taking advantage of user movement to refine measured 3D bounding boxes, lightweight by avoiding heavyweight 3D object detection and resorting to the mature 2D counterpart, and hybrid by effectively splitting workload between the edge (\ie RGB-image-based 2D object detection) and the headset (\ie depth-image-based 3D bounding box estimation).
We depict the system architecture of \name in Figure~\ref{fig:architecture}.

The design of \name is inspired by three key observations of existing solutions for object detection and the differences between smartphone-based and headset-based mobile AR/MR.
First, while DNN-based 3D object detection leads to high computation latency even when assisted by the edge, its 2D counterpart is computation-efficient (\eg 13 ms latency in Figure~\ref{fig:latency}).
Second, although mobile headsets are equipped with various sensors to facilitate AR/MR applications, their hardware resources are typically constrained and the low-resolution depth images limit the performance of 3D object detection models.
Third, headset-based AR/MR differs from smartphone-based one by rendering the bounding box using the physical location of the object, instead of its relative position in the camera view.

The overarching goal of \name is to 
simultaneously reduce end-to-end latency and increase detection accuracy, improving QoE for next generation headset-based AR/MR.
To achieve the above goal, we face the following challenges when designing \name.

\BULLET How to jointly consider existing techniques to reduce end-to-end latency of 3D object detection?

\BULLET How to accurately and efficiently measure/estimate the 3D bounding box of an object from depth data on the headset?

\BULLET How to boost the performance of \name under different scenarios for improving user experience?

Next, we present how we address these challenges.
\mysection {System Design of \bigname}
\label{sec:design}

In this section, we introduce the basic design of \name. 
We will explain how to improve its performance in \S\ref{sec:optimization}.

\mysubsection{Edge-assisted 2D Object Detection}
\label{sec:2d}

As shown in Figure~\ref{fig:architecture}, the workflow of \name begins with retrieving RGB images and offloading them to the edge for conducting 2D object detection.
Given that \name is a generic framework, it can work with any DNN-based model that can accurately label objects and generate their 2D bounding boxes in real time~\cite{DBLP:journals/corr/abs-2004-10934, Girshick2015FastR, Redmon2018YOLOv3AI}.
The 2D bounding box drawn on the RGB image will be used as the starting point to derive the 3D bounding box using depth data.
Since the main purpose of the 2D bounding box is to reduce the amount of to-be-processed depth data, the key requirement is that the object should completely fit into the returned bounding box, which could be larger than the object if doing this can speed up 2D object detection.
We will describe how to optimize the offloading efficiency in terms of data usage in \S\ref{sec:RoI}.

\mysubsection{Bounding Box Alignment on Depth Frame}
\label{sec:alignment}

\begin{figure}
  \centering
    \begin{minipage}{.56\linewidth}
      \centering
       \includegraphics[width=0.9\linewidth]{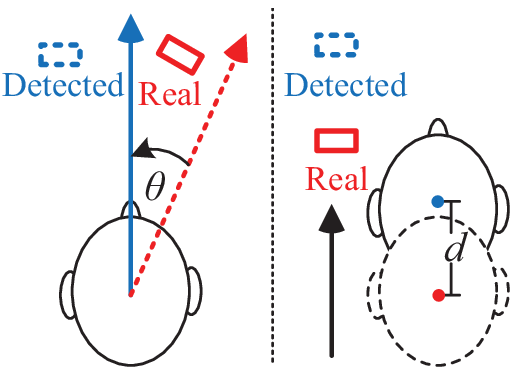}
       \vspace{-0.1in}
       \caption{\small Misalignment caused \\by  
       movement.}
       \label{fig:error_compensation}
       \vspace{-0.2in}
    \end{minipage}%
    \begin{minipage}{.42\linewidth}
      \centering
      \vspace{0.20in}
      \includegraphics[width=1.05\linewidth]{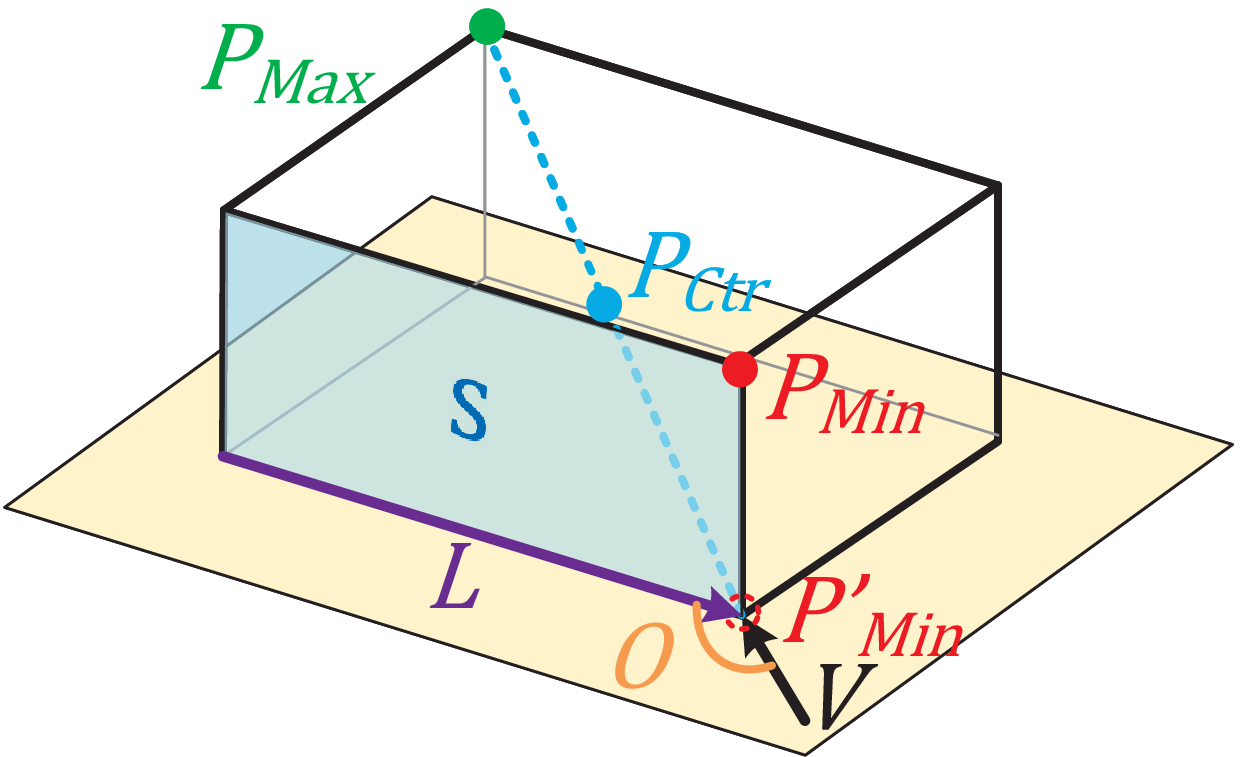}
	    \vspace{-0.20in}
	   	\caption{\small Geometric model \\of bounding box.}
	    \label{fig:rdml}
	    \vspace{-0.2in}
    \end{minipage}%
\end{figure}

The next step is to align 2D bounding boxes from the RGB image onto the depth image.
Since it takes time to get the results from the edge, during which the camera view may change due to movement, we need to first transform the returned 2D bounding boxes on the offloaded image to the current viewport.
Otherwise, there will be a misalignment between 2D bounding boxes and objects, as shown in Figure~\ref{fig:error_compensation}.
To solve this problem, \name records the 6DoF pose of the frame when it is captured by the RGB camera, which is provided by the headset.
Once users start the headset, its motion sensors (\eg gyroscope, accelerometer, visible light cameras, \etc) begin to track the headset's 6DoF pose during movement and make it available to applications. 
After receiving the detection results from the edge, it can transform the 2D bounding box to the current viewport based on the change of 6DoF pose (\ie $\theta$ and $d$ in Figure~\ref{fig:error_compensation})~\cite{roetenberg2009xsens}.

\name then calculates the coordinate of the center pixel for a detected object using the updated 2D bounding box, based on its four vertices ($R_{Right}$, $R_{Left}$, $R_{Top}$, $R_{Bottom}$), as $R_{Ctr} = ((R_{Right}+R_{Left})/{2}, (R_{Top}+R_{Bottom})/{2})$.
Different from the setup of RGB-D cameras, most headsets are equipped with an RGB camera and a depth camera that are not synchronized with each other.
As a result, both the center point and the resolution of the depth frame are different from those of the corresponding RGB image (captured at the same time). 
%
%
To determine the center point $P'_{Ctr}$ on depth frame that is mapped to the center on RGB frame, we can utilize the pinhole camera principle~\cite{strobl2011more}. 
%
%
%
Note that $P'_{Ctr}$ is just a point on the surface of the object on the depth image, not the actual center point of the 3D bounding box.
This point will be used for determining one of the surfaces of the 3D bounding box (\S\ref{sec:dimension}).
Similarly, we can get the corresponding points on the depth frame for the four vertices 
of the 2D bounding box, which will be used for background removal (\S\ref{sec:removal}).

\mysubsection{Background Removal on Depth Image}
\label{sec:removal}

After getting the bounding box of the object on the depth image, in this step, \name removes the background in the bounding box to reduce computation overhead and improve the accuracy of our 3D bounding box estimation, by leveraging existing solutions developed by the computer graphics (CG)/CV communities~\cite{Floor}.
An alternative solution is to perform semantic segmentation, which can label each pixel, instead of object detection on the edge.
However, this will increase both data transmission overhead by sending per-pixel labels and computation overhead to match each pixel of the object onto the depth image.
After removing the background, we can obtain depth data of mainly the detected object. 
Since the collected depth data may contain noises and undetected areas of the object, the depth information of the above vertices and the center point on the depth frame may be missing. 
To improve the quality of depth frames, we further apply an edge-preserving filter and Spatial Hole-filling algorithm~\cite{grunnet2018depth} to the depth frame to make it smooth and complete.

\mysubsection{Center Point Estimation}
\label{sec:center}

There are three parameters to determine the 3D bounding box of an object, the spatial position (\ie location), 3D dimensional size (\ie height, width, and thickness), and orientation. 
In this section, we estimate the center point of the 3D bounding box (\ie the spatial position of the object).
We measure the dimension and orientation of the object in \S\ref{sec:dimension}.
%
%
With the depth data, \name can get the spatial coordinates of the closest point $P_{Min}$ and the furthest point $P_{Max}$ 
to the headset, as shown in Figure~\ref{fig:rdml}. 
After that, it projects $P_{Min}$ to the plane that the object is placed on, which could be detected by the headset, to get  $P'_{Min}$. 
The center point of the 3D bounding box $P_{Ctr}$ is estimated as the center of the line connecting $P'_{Min}$ and $P_{Max}$.
%
The estimation of the two key points may not always be accurate, especially for $P_{Max}$, as the actual furthest point may be occluded and thus not visible.
By exploiting the headset movement, we propose an optimization to further improve the accuracy when users move around the object to observe more details (\S\ref{sec:refine}), which is a typical use case for headset-based mobile AR/MR.

\mysubsection{Dimension and Orientation Estimation}
\label{sec:dimension}

To get the dimension and orientation of the 3D bounding box, \name first detects the surface $S$ that $P'_{Ctr}$ is on, as shown in Figure~\ref{fig:rdml}\footnote{$S$ could be the other vertical surface shown in Figure~\ref{fig:rdml}, but this does not affect the estimation (width vs. thickness). $S$ could also be the top horizontal surface. In this case, \name keeps moving $P'_{Ctr}$ toward a direction until it hits the ground.}, with the following method.
It uses $P'_{Ctr}$ as a start pixel of a seed patch~\cite{fischler1981random} on the depth frame. 
It then grows the patch to a certain size and utilizes the linear least-squares plane fitting~\cite{lukacs1998faithful} to identify the best fitting plane for this patch.
This plane will be used to approximate the surface $S$ in Figure~\ref{fig:rdml}.
To improve the accuracy of this estimation, \name can repeat the above process multiple times with different start pixel of the seed patch, for example, by using other points close to $P'_{Ctr}$, and then aggregate the calculated planes to approximate $S$.

With the surfaces $S$ and the center point $P_{Ctr}$ (\S\ref{sec:center}), \name can calculate the dimension of the 3D bounding box.
If the distance between $P_{Ctr}$ and the underlying plane is $d_h$, the object height $H$ is $2 \times d_h$. 
%
Next, \name calculates the distance $d_t$ between $P_{Ctr}$ and $S$.
The thickness of the bounding box $T$ will be $2 \times d_t$.
With $H$ and $T$, we can calculate the width of the bounding box $W$ by using the right angle theorem: $W = \sqrt{d_{P'P}^2 - H^2 - T^2}$, where $d_{P'P}$ is the distance between $P'_{Min}$ and $P_{Max}$.
%
%
After getting the spatial position and dimension of the object, \name still needs to determine the orientation $O$ of the 3D bounding box.
It first calculates the intersecting line $L$ of the surface $S$ and underlying plane.
From the 6DoF pose, \name knows the viewing direction $V$ of the user $D$. 
Thus, $O$ can be calculated based on the angle between $L$ and $V$, as shown in Figure~\ref{fig:rdml}.

\mysection{Performance Optimizations} 
\label{sec:optimization}

\mysubsection{Motion-aware Bounding Box Refinement}
\label{sec:refine}

A unique feature of \name is that it can keep refining estimated 3D bounding boxes when users move around an object of interest, for example, to investigate the details.
%
%
As we will show in \S\ref{sec:accuracy-mobile}, existing DNN-based 3D object detection models cannot benefit from headset movement in their current form.
This refinement mode is enabled only when users move around an object, which can be inferred from the 6DoF pose of the headset and the location of the object.
For two consecutive 3D bounding boxes that are estimated by \name, it first gets the spatial point that is the center of the line connecting the two center points of the two boxes.
It then uses this point as the center of the updated bounding box and moves the two estimated boxes to this point.
It finally uses the union of the two boxes as the updated box, which will be combined with the next estimated bounding box.

A key difference between video see-through based AR/MR on smartphones and optical see-through based one on headsets is that the latter does not need to continuously offload camera views to the edge even when users move.
The location of an object displayed on the screen of smartphones changes if users move, which requires conducting object detection on the updated camera view.
Optical flow tracking can alleviate this issue only to some extent, as the tracking error will accumulate as time goes on.
On the other hand, the 3D bounding box of an object is determined by its actual physical location and orientation that will not change with headset movement.
The underlying coordinate-system drift caused by movement will be fixed by the headset itself, and thus \name can always get an accurate pose from the headset to update the rendered bounding box.
As a result, headset-based AR/MR does not need to frequently perform (edge-assisted) object detection.
To optimize the overhead of \name's bounding box refinement, we next introduce the motion-assisted dynamic region of interest (RoI) encoding to optimize the offloading overhead.

\mysubsection{Motion-assisted Dynamic RoI Encoding}
\label{sec:RoI}

Dynamic RoI encoding selectively applies lossy compression to parts of the frame that are less likely to contain objects of interest and lossless compression to other areas for reducing the amount of encoded data.
The RoIs on the current frame are determined by analyzing the microblocks of 2D images and checking whether they overlap with the identified RoIs in a previous frame.
This scheme has been demonstrated to be helpful for AR on handheld smartphones with limited moving speed~\cite{Liu:2019:EAR}.
However, the camera view of the headset may drastically change with user movement. 
For example, the peak speed of head movement can reach 240 degrees per second~\cite{fang2015eye}, much higher than the moving speed of a smartphone when used for AR and making microblock-based scheme less effective for headsets. 

\name resorts to the 6DoF tracking offered by headsets to solve this problem.
By recording the 6DoF pose of consecutive frames, \name can determine whether they overlap with each other.
If not, dynamic RoI encoding will not be applied.
Otherwise, \name checks whether there are known RoIs of a previous frame appearing on the current frame and (if they do) get their locations on the current frame through coordinate transformation.
\name compresses the identified RoIs and the area that is not overlapped with the previous frame losslessly.
It compresses the remaining area in a lossy fashion.
Note that dynamic RoI encoding is a generic design and can be applied to not only the bounding box refinement mode but also other scenarios.

\mysubsection{3D Bounding Box Caching and Reusing}
\label{sec:caching}

To better support mobile scenarios where users move around to explore the surrounding environment, we design a mechanism to cache and reuse 3D bounding boxes of detected objects, which avoids unnecessary detection of the same object multiple times. 
The goal is to display the 3D bounding box of a detected object as fast as possible, when it reappears, by reducing the initial rendering time, which can boost user experience and decrease computation resource utilization on both the edge and the headset.
This optimization is helpful, especially under dynamic network conditions that increase end-to-end latency of object detection.

In the cache, we store the 6DoF pose and 3D dimension of detected objects.
When users move, \name keeps updating the viewing frustum (\ie 3D viewport) based on the 6DoF pose of the headset and checks whether there are cached items that should be in the current viewport by examining the 6DoF pose of cached bounding boxes.
To further reduce the rendering time of 3D bounding boxes for cached items, \name saves their translucent cubes in the memory. 
Based on the cached results, it can reshape and rotate the cubes and immediately render them on display. 
After that, \name performs object detection on the current viewport, in case there are new objects in the scene, and updates the cache accordingly. 
Another benefit of our caching design is that if a cached item is further away from the user in the updated viewport and out of the range of the depth camera, \name can still render its bounding box retrieved from the cache. 

\mysection{System Implementation} 

We develop a prototype implementation of the \name client on Microsoft HoloLens 2 and the \name edge server on Linux.
We implement the device-side functions with Windows SDK~\cite{WinSDK}, DirectX~\cite{DirectX}, and Unity 3D engine~\cite{unity}.
We use multi-threading to simultaneously read data from both RGB and depth cameras. 
We collect the camera frame using libraries of Windows SDK. 
When receiving the 2D detection results from the edge server, we obtain the depth frame by enabling the Research Mode of HoloLens. 
We store depth images in bitmaps to improve the speed of data processing.
To enable motion-based dynamic RoI encoding, we utilize the position and orientation of the headset, which are retrieved via a library in DirectX.
After estimating the 3D bounding box, we render it on the screen with Unity.
By adapting the detection results from the previous frame to the change of users' 6DoF pose position, we encode the RoI of the current frame using JPEG and send it to the edge. 
We implement the \name edge server in Darknet~\cite{darknet13} open-source neural networks.
The edge provides 2D object detection for \name using YOLOv4~\cite{DBLP:journals/corr/abs-2004-10934}.
As a generic framework for 3D object detection, \name can use any mature 2D object detection model on the edge.

In total, our implementation consists of 4,600+
lines of code (LoC): 3,000+ LoC in C\# (rendering, device localization, and bounding box estimation) and 1,000+ LoC in C++ (gathering sensor data, image compression, and networking) for the client, and 600+ LoC in C++ (networking and multi-threading) for the server.
We also build a prototype of \name on HoloLens (1st gen), on which the performance of \name is only slightly worse than that of HoloLens 2.
Hence, we report the results for only HoloLens 2.

\mysection{Performance Evaluation} 
\label{sec:results}

In this section, we measure the performance of \name through dataset-driven evaluations, controlled (live) experiments, and an IRB-approved user study.


\mysubsection{Experimental Setup}
\label{sec:setup}

We compare the performance of \name with the following {\em eight} start-of-the-art 3D object detection models, COG~\cite{ren2016three}, VoteNet~\cite{qi2019deep}, and MLCVNet~\cite{DBLP:conf/cvpr/XieLWWZXW20} with point clouds as input, F-PointNet~\cite{qi2017frustum} and Trans3D~\cite{DBLP:conf/iccv/TangL19} for models using RGB-D input, and  Mono3D~\cite{DBLP:conf/cvpr/ChenKZMFU16}, AM3D~\cite{DBLP:conf/iccv/MaWLZOF19}, and D$^4$LCN~\cite{DBLP:conf/cvpr/DingHYWSLL20} that take 2D images as input.

\begin{figure}[t]
\centering
\includegraphics[width=0.8\linewidth]{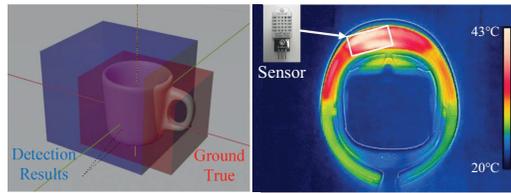}
\vspace{-0.15in}
\caption{
\small Visualization of 3D IoU (left) and position of external temperature sensor (right).}
\label{fig:boundingbox}
\vspace{-0.2in}
\setlength{\belowcaptionskip}{-1cm}
\end{figure}

\vspace{0.05in}
\noindent{\bf Testbed.}
The edge server is equipped with an Intel i9-9900k CPU, an NVIDIA RTX 2080S GPU, and 64GB DDR4 3200MHz RAM. 
The headset, Microsoft HoloLens 2, and the edge run the Universal Windows Platform (version 10.0.20346.0) and Ubuntu 16.04, respectively.
For most experiments, we connect the headset and the edge 
with a Linksys AC1900 WiFi router that is attached to the same 1 Gbps Ethernet as the edge server. 
The normal throughput of this WiFi network is around 260 Mbps, and its round trip delay is less than 1 ms.
We use this WiFi router exclusively for our experiments, by avoiding interference with other co-existing WiFi networks.
For the experiments under dynamic network conditions, we attach an LTE modem to the headset, which connects to the edge server through our USRP-based LTE base station.
The throughput of this LTE network ranges from 8.4 to 37.1 Mbps, and its typical round trip delay is about 14 ms.

\vspace{0.05in}
\noindent{\bf Evaluation Metrics.}
We use the accuracy of 3D object detection and the end-to-end latency as the key metrics to evaluate \name.
We also measure the surface temperature, battery power level, and other computation resource utilization (\eg CPU, GPU, and memory) on Microsoft HoloLens 2.

{\em 3D Intersection over Union.}
We evaluate the accuracy of the 3D bounding box using 3D Intersection over Union (3D IoU), as shown in Figure~\ref{fig:boundingbox} (left), which has been widely used in the literature~\cite{Chen_2017_CVPR,ku2018joint,Liu_2019_CVPR,DBLP:conf/cvpr/QiSMG17,shi2020pvrcnn,Shi_2019_CVPR,Wang_2021_CVPR}. 
%
%
%
By following the common practice in the computer vision community~\cite{Chen_2017_CVPR,ku2018joint,DBLP:conf/cvpr/QiSMG17,Shi_2019_CVPR,Wang_2021_CVPR}, we set the 3D IoU thresholds to be 0.25 and 0.5, respectively. 
That is to say, when the 3D IoU is larger than the threshold, we consider the detection result to be accurate. 
In the following, we report the percentage of accurate detections using the 3D IoU metric.

{\em End-to-end Latency.} The end-to-end latency is important for real-time, interactive AR/MR systems. 
%
We record the time $t_i$ when the $i$th frame is captured by the camera and the time $\hat{t}_i$ when the 3D bounding boxes are rendered for it. 
The latency of the $i$th frame is defined as $\tau_i = \hat{t}_i - t_i$. 
Let $n$ be the number of processed frames. 
The end-to-end latency can be expressed as $\Delta=\sum_{i=1}^{n}\tau_i/n$.

{\em Surface Temperature.} 
The headset's surface temperature is measured by a sensor, as illustrated in Fig.~\ref{fig:boundingbox} (right). 
Using a FLIR E8-XT infrared camera~\cite{Heat}, we capture the thermal image of the headset after 20-minute use. 
Based on the thermal image, we locate the area where the headset generates the most heat and place an LM35 temperature sensor~\cite{Sensor} there. 
The sensor records and transmits the temperature data every second to the headset over WiFi. 

{\em Battery Power Level.} To monitor and analyze the battery power level of the headset, we disassemble it and remove its battery. 
In the experiment, a Keithley 2281S battery simulator~\cite{Battery} is used as the power supply, which can provide the headset with a stable DC input and monitor the current and power. 
In this way, we can have a systematic evaluation of the headset's battery power level.

\vspace{0.05in}
\noindent{\bf Evaluation Datasets.}
We train the state-of-the-art 3D object detection models with the well-known SUN RGB-D dataset~\cite{song2015sun}, and train YOLOv4~\cite{DBLP:journals/corr/abs-2004-10934} that is used in \name for 2D object detection with the widely-used COCO~\cite{lin2014microsoft} dataset. 
One unique challenge of comparing the performance of \name with existing models is it needs extra information that is not included in regular RGB-D datasets such as SUN RGB-D.
The reason is that to make \name lightweight and suitable for mobile headsets, we leverage the 6DoF pose of the headset to estimate the 3D bounding box, which is missing from existing RGB-D datasets.

To address this challenge and conduct a fair comparison of \name and existing models, we construct a dataset that consists of not only RGB-D images but also the 6DoF pose of the camera (\ie an Intel RealSense D435~\cite{D435} camera) when capturing the images.
Our dataset has 2,184 RGB-D images captured with different viewing angles and distances, as shown in Figure~\ref{fig:setting}.
The RGB-D images also have different resolutions, which is the reason that we choose Intel RealSense, instead of HoloLens that generates low-resolution depth images, as the capturing device.
This RGB-D dataset contains seven classes of objects, table, chair, bottle, box, desk, bag, and TV, which are common objects in the public datasets~\cite{lin2014microsoft,song2015sun}.
Our dataset is diverse as it covers objects of different sizes, shapes, and materials (which affect the generation of depth images), and some objects (\eg chair, bottle, and bag) may have irregular shapes.
The whole dataset is annotated with the ground truth 3D bounding boxes (\ie accurate object orientation and position).  
We plan to make our collected dataset publicly available, which hopefully can benefit future research on designing more efficient and accurate 3D object detection models for mobile headsets.

\begin{table*}[!htp]
\centering
\small 
\begin{tabular}{c|c|ccccccc|c}

\multirow{2}{4em}{Methods} &\multirow{2}{2.3em}{Input} & & \multicolumn{5}{c}{3D IoU@0.25/0.5} &  & \multirow{2}{3.5em}{Average} \\ \cline{3-9}   
 &  & \centering Table &\centering Chair &\centering Bottle &\centering Box &\centering Desk &\centering Bag &\centering TV &  \\    \hline 

COG~\cite{ren2016three}       & PC  & 47.3/2.6 & 39.3/- & 46.4/1.8 & 57.3/15.2  & 49.5/13.3 & 45.1/10.3 & 39.5/5.7 & 47.9/7.2                             
\\ 

VoteNet~\cite{qi2019deep}   & PC  & 56.2/23.6 & 45.5/22.2 & 67.2/18.3 & 52.4/22.6  & 48.6/14.2 & 54.3/18.5 & 41.9/19.2 & 51.6/21.7
\\ 
MLCVNet~\cite{DBLP:conf/cvpr/XieLWWZXW20} & PC  & 61.3/22.5 & \textbf{74.2}/27.6 & \textbf{68.2}/16.4 & 62.1/22.3  & 63.7/26.5 & 57.6/21.1 & 52.3/17.9 & 63.1/22.1                                 
\\
F-PointNet~\cite{qi2017frustum}   & RGB-D & 45.5/11.2 & 37.5/17.3 & 44.2/9.5 & 31.6/17.5  & 38.5/- & 54.5/17.6 & 37.4/8.2 & 42.3/11.9
\\

Trans3D~\cite{DBLP:conf/iccv/TangL19} & RGB-D & 52.3/13.5 & 38.4/- & 47.2/13.2 & 39.5/8.4 & 47.2/18.4 & 49.1/- & 37.2/12.9 & 45.8/10.1
\\

Mono3D~\cite{DBLP:conf/cvpr/ChenKZMFU16} & RGB & 59.3/11.2 & 52.4/9.5 & 47.3/8.6 & 65.3/13.2 & 67.5/21.5 & 48.6/5.2 & 54.5/16.4 & 55.4/10.7
\\

AM3D~\cite{DBLP:conf/iccv/MaWLZOF19}    & RGB & 65.9/14.9 & 71.2/11.3 & 62.7/16.3 & 57.8/10.7 & 49.2/12.5 & 51.7/11.2 & 57.9/12.9 & 58.7/13.6
\\
D$^4$LCN~\cite{DBLP:conf/cvpr/DingHYWSLL20} & RGB & 68.3/14.3 &  69.2/24.4 & 67.2/18.6  & 58.9/21.5 & 62.3/14.3 & 56.6/9.6 & 57.2/11.7 & 63.9 /15.9
\\

\hline
DeepMix & Hybrid & \textbf{72.4/37.5} & 67.4/\textbf{33.7} & 63.1/\textbf{38.4} & \textbf{66.5/39.1}  & \textbf{72.1/33.2} & \textbf{65.8/46.5} & \textbf{61.2/31.8} & \textbf{67.4/37.2}
\\
\end{tabular}
\caption{\small 
Class-wise 3D IoU@0.25/0.5 comparison of \name and state-of-the-art 3D object detection models. }
\label{tab:3DIoU-static}
\end{table*}

\mysubsection{End-to-end Latency}
\label{sec:latency}

We first compare the end-to-end latency of \name with existing 3D object detection models by replaying the RGB-D images in our collected dataset on HoloLens 2 and plot the experimental results in Figure~\ref{fig:e2e}.

We break down the end-to-end latency into four parts, data transmission, point cloud construction, server-side data processing, and client-side data processing.
\name and image-based models send only RGB images to the edge; whereas the others send RGB-D images to the edge.
Note that instead of generating point clouds on the headset and sending them to the edge for object detection, we send RGB-D images and create point clouds on the edge.
The reason is that doing this can not only reduce computation resource utilization, energy consumption, and surface temperature of the headset (\S\ref{sec:resource}), but also drastically save network data usage.
For example, the size of a point cloud generated from an RGB image and a depth image (with a combined size of 1.68 MB) could be as large as 35.9 MB. 
The server-side data processing latency is mainly the inference time of DNN models for 2D/3D object detection.
For \name, the estimation of 3D bounding boxes happens on the headset, after receiving the returned 2D bounding box.
The client-side data processing of existing models mainly includes the transformation of returned 3D bounding boxes to the coordinate system of the headset for rendering and display.

The key observation from Figure~\ref{fig:e2e} is that the end-to-end latency of \name is significantly lower than existing 3D object detection models.
It takes only 34 ms for \name to accurately detect 3D objects, among which the latency is 5 ms (14.7\%) for data transmission, 13 ms (38.2\%) for server-side data processing, and 16 ms (47.1\%) for client-side data processing, respectively.
Compared to image-based models, \name's dynamic RoI encoding scheme (\S\ref{sec:RoI}) effectively reduces the data transmission time by 34\% (from 6.7 ms to 5 ms).
As \name estimates the 3D bounding box on the headset, its client-side data processing time (16 ms) is slightly longer than that of other schemes (13.8 ms for image-based models and models with RGB-D input, 11 ms for point-cloud-based models).
Thus, thanks to its lightweight design (\S\ref{sec:design}) and various optimizations (\S\ref{sec:optimization}), the latency of \name is far below the 100 ms requirement for interactive AR/MR.

\begin{figure*}
  \centering
    \begin{minipage}{.24\linewidth}
      \centering
      \vspace{-0.2in} 
      \includegraphics[width=0.99\linewidth]{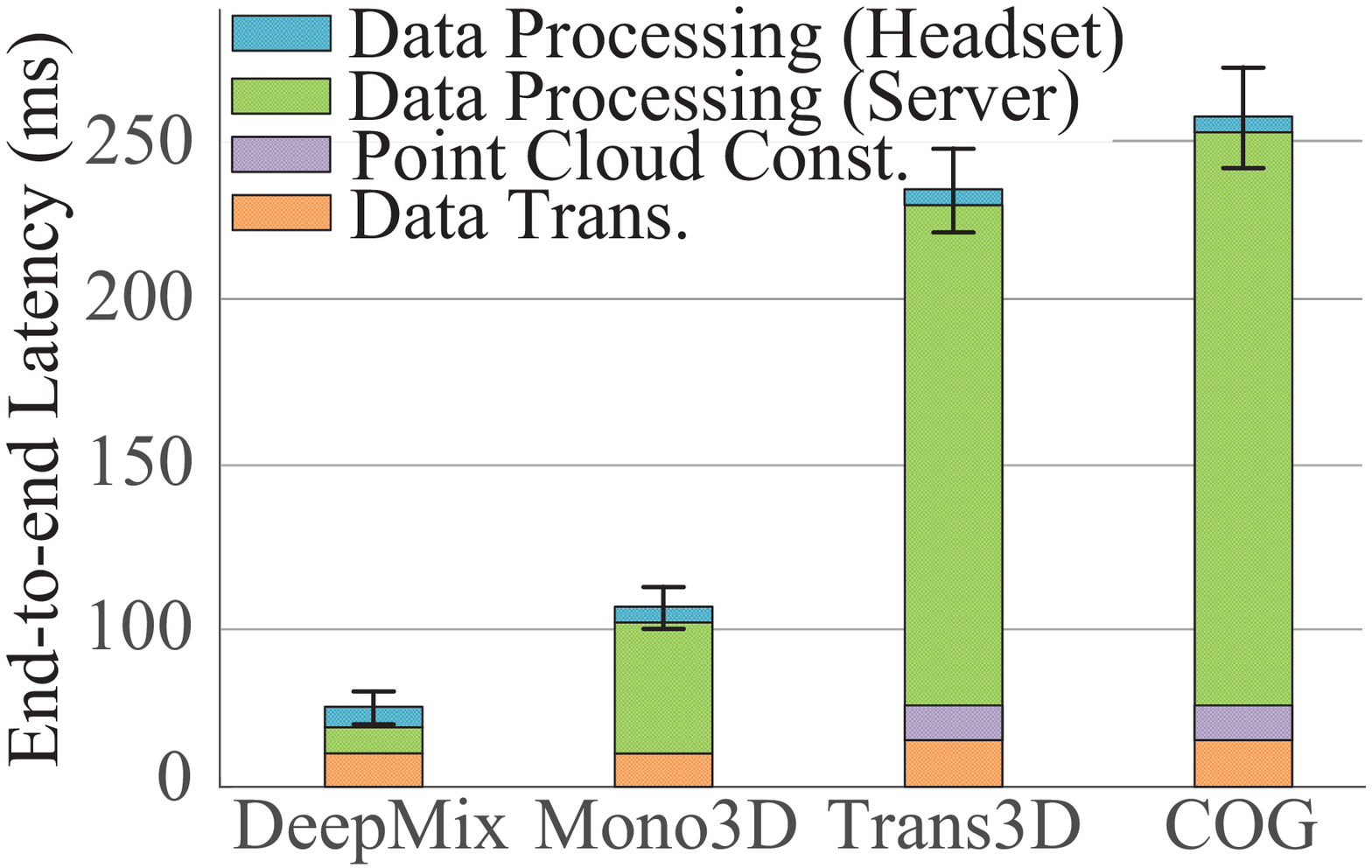}
      \vspace{-0.27in}
      \caption{\small End-to-end \\Latency on LTE Network.}
      \label{fig:LTE}
      \vspace{-0.25in}    
    \end{minipage}%
    \begin{minipage}{.46\linewidth}
      \centering
      \includegraphics[width=0.9\linewidth]{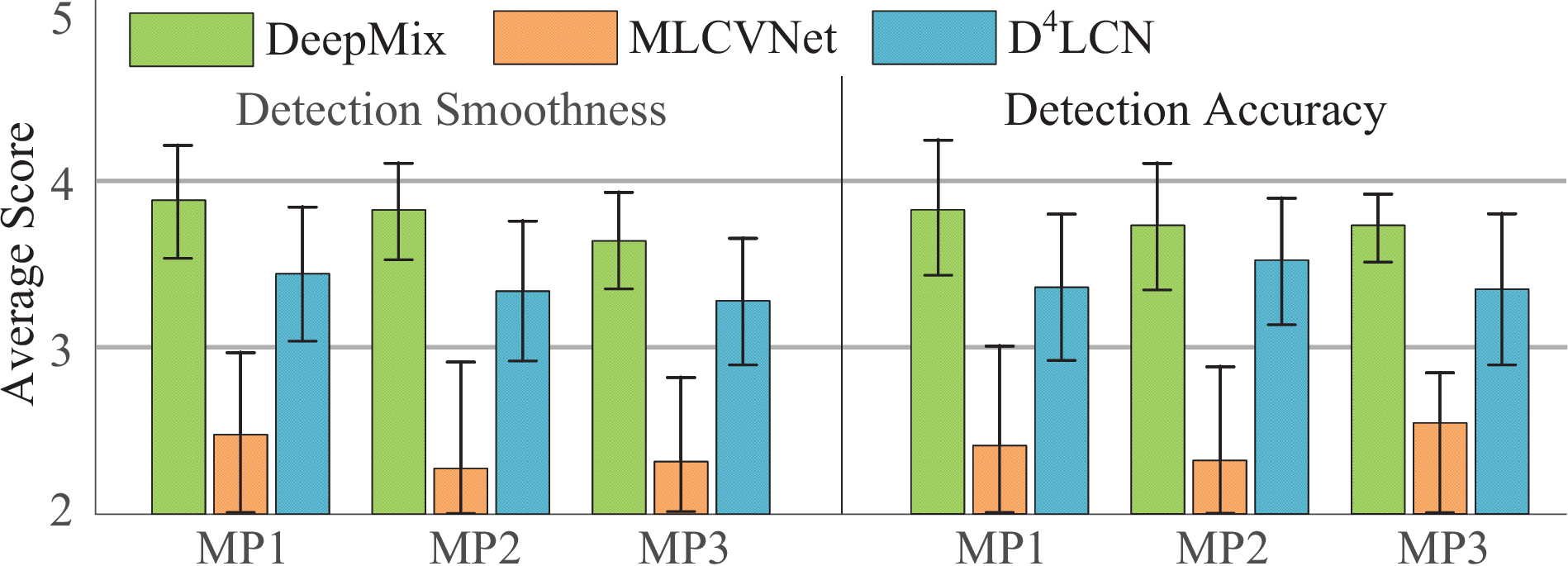}
      \vspace{-0.1in}
      \caption{\small Evaluation of the smoothness and accuracy of 3D \\object detection through a user study with 33 participants. 
      }
      \label{fig:user-study}
      \vspace{-0.20in}
    \end{minipage}%
    \begin{minipage}{.23\linewidth}
      \centering
      \includegraphics[width=0.99\linewidth]{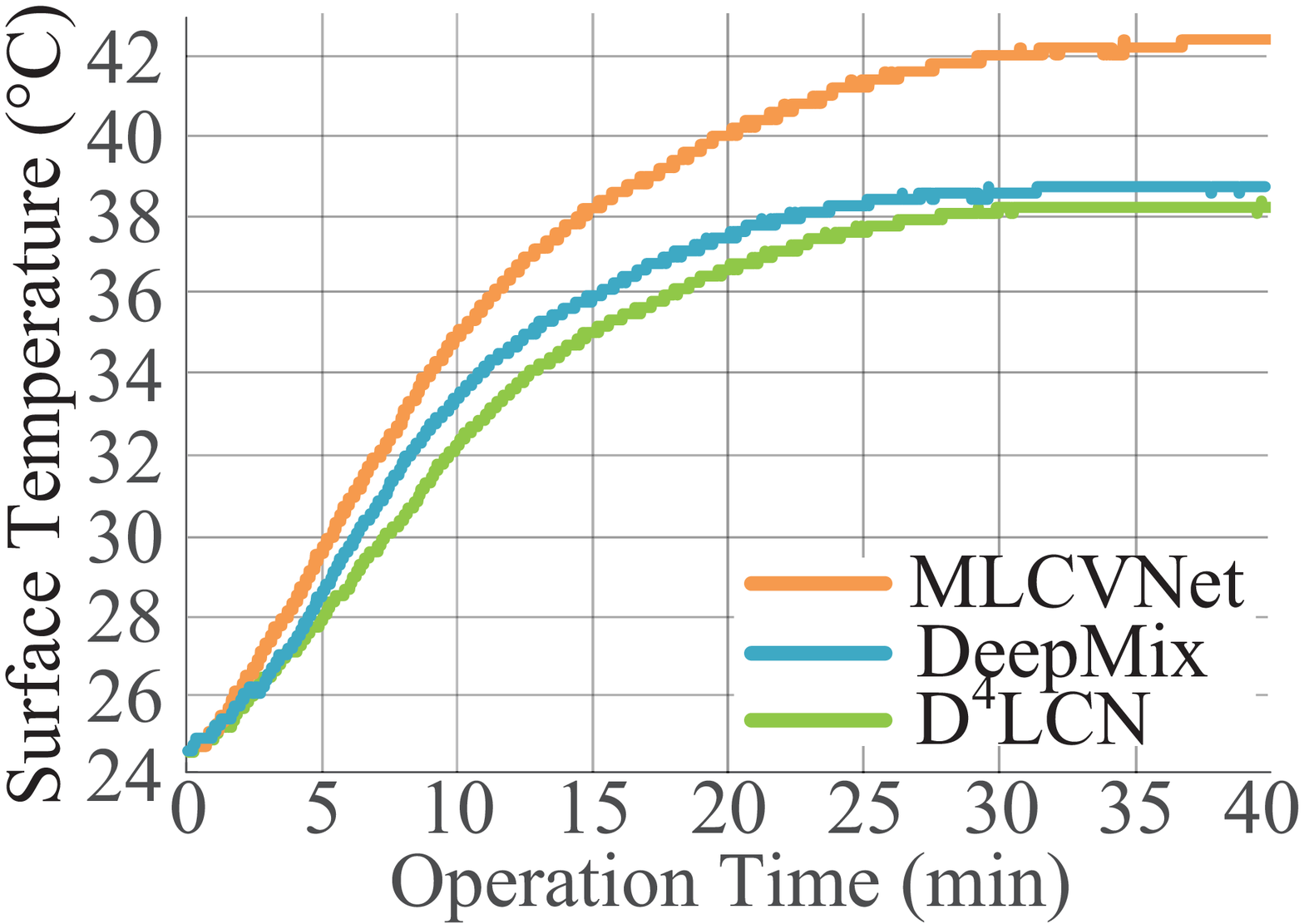}
      \vspace{-0.3in}
      \caption{\small Surface Temperature.}
      \label{fig:temp}
      \vspace{-0.25in}
    \end{minipage}%
\end{figure*}

For existing 3D object detection models, their end-to-end latency is dominated by the server-side processing time, especially for point-cloud-based solutions.
Even on the edge server, it takes about 23 ms to generate the point clouds.
Note that although the end-to-end latency of Mono3D~\cite{DBLP:conf/cvpr/ChenKZMFU16} is also lower than 100 ms, as we will show next (\S\ref{sec:accuracy-static} \& \S\ref{sec:accuracy-mobile}), its detection accuracy is much lower than that of \name, and its latency will be higher than 100 ms under dynamic network conditions (Figure~\ref{fig:LTE}).
Figure~\ref{fig:e2e} shows that the data transmission time of existing models with RGB-D and point cloud inputs are higher than that of \name and image-based models, due to the additional depth images that are needed to learn the 3D bounding box.
We conduct the above experiments on the WiFi network with high throughput (\S\ref{sec:setup}).
When the network condition becomes worse, this data transmission time of RGB-D data will be more noticeable.
We will evaluate the end-to-end latency on an LTE testbed in \S\ref{sec:network}.

\mysubsection{Detection Accuracy in Static Scenario}
\label{sec:accuracy-static}

We compare the detection accuracy of \name with existing models for the static scenario using the 3D IoU metric (\S\ref{sec:setup}).
To make the comparison fair, we replay on the headset the 2,184 RGB-D images in our collected dataset (\S\ref{sec:setup}).
%
%
%
We present the results for 3D IoU in Table~\ref{tab:3DIoU-static}.
For the static scenario, remarkably, \name achieves better overall performance than all other models across different object categories. 
In this table, the `-' symbol means that the corresponding method could not detect the object. 
On average, \name has the highest 3D IoU for both the 0.25 and 0.5 thresholds.
One possible reason is that instead of completely relying on learning-based models, \name estimates the 3D bounding box using pixel-level depth information on the headset.

When the threshold is 0.5, \name's accuracy is the highest for all object classes, and is 1.68$\times$ (on average) over the best existing model MLCVNet~\cite{DBLP:conf/cvpr/XieLWWZXW20}.
When the threshold changes to 0.25, the detection accuracy of \name is still higher than the best state-of-the-art model D$^4$LCN~\cite{DBLP:conf/cvpr/DingHYWSLL20} by 3.5\% (on average).
In this case, MLCVNet~\cite{DBLP:conf/cvpr/XieLWWZXW20} performs the best for chairs and bottles.
The reason is that the depth data will be missing when there is a reflection on the bottle, affecting the accuracy of \name.
We are exploring efficient computer graphics algorithms~\cite{ji2017fusing, sajjan2020clear} to address this problem\footnote{The scheme~\cite{grunnet2018depth} in \name is {\em lightweight} and cannot handle it.}. 
For chairs with irregular shapes, 
the 3D bounding box estimation of \name may not be accurate from some specific viewing angles.
This issue could be alleviated in mobile scenarios (\S\ref{sec:accuracy-mobile}). For example, if users are interested in an object, they may move around it to inspect the details from different angles, which offers opportunities to improve detection accuracy (\S\ref{sec:refine}).
%
\name still outperforms 5 out 8 existing models and has the most accurate results when the threshold is 0.5, 1.22$\times$ (on average) over the best existing model MLCVNet~\cite{DBLP:conf/cvpr/XieLWWZXW20} (33.7\% vs. 27.6\%), for this challenging ``chair'' category.
We also evaluate the detection accuracy with another widely-used metric, called mean spatial position accuracy (mSPA)~\cite{FRISOLI20111929}.
The results (not shown due to the limited space) are qualitatively similar to those of 3D IoU.

\mysubsection{Detection Accuracy in Mobile Scenarios}
\label{sec:accuracy-mobile}

\begin{table*}[!htp]
\centering
\small 
\begin{tabular}{c|c|cccc|c}

\multirow{2}{4em}{Methods} & \multirow{2}{2.3em}{Input} & \multicolumn{4}{c|}{3D IoU@0.25 (\hd{Scenario 3}: speed@0.5/1/2m/s)} & \multirow{2}{3.5em}{Average} \\ \cline{3-6}   
 & &\centering Chair  &\centering Bottle &\centering Box  &\centering Bag  &  \\    \hline 

COG~\cite{ren2016three}     & PC  &  38.5/33.2/27.4 &  32.2/27.6/19.7 &  46.2/34.1/28.6  & 36.8/28.8/21.6 & 38.4/30.9/24.3                             
\\ 

VoteNet~\cite{qi2019deep}   & PC  &  48.4/35.5/29.2 & 52.1/43.2/24.6  & 48.5/36.7/28.2  & 45.2/30.2/18.1  & 48.6/36.4/25.1
\\ 
MLCVNet~\cite{DBLP:conf/cvpr/XieLWWZXW20} & PC  &  55.7/42.2/28.4 & 53.7/43.2/33.9  & 45.2/30.2/19.8 & 44.6/25.1/17.2 & 49.8/35.1/24.8                                 
\\
F-PointNet~\cite{qi2017frustum}   & RGB-D &  32.8/21.3/17.6 & 35.2/25.9/21.2  & 25.6/14.1/8.9  & 41.8/32.4/19.6  & 33.9/23.4/16.8
\\

Trans3D~\cite{DBLP:conf/iccv/TangL19} & RGB-D &  33.8/25.7/22.4 & 34.6/22.5/17.6  & 33.7/25.4/15.5  & 45.5/36.2/18.6  & 36.9/27.5/18.5
\\

Mono3D~\cite{DBLP:conf/cvpr/ChenKZMFU16} & RGB &  59.6/49.1/41.6 & 45.2/37.6/23.9  & 66.5/51.1/36.2  & 41.7/34.1/27.7  & 53.2/42.9/32.3
\\

AM3D~\cite{DBLP:conf/iccv/MaWLZOF19}    & RGB &  64.5/54.2/47.3 & 59.4/47.2/32.9  & 63.3/54.9/46.2  & 61.4/53.2/42.5  & 62.1/52.4/42.2
\\
D$^4$LCN~\cite{DBLP:conf/cvpr/DingHYWSLL20} & RGB &   65.3/49.2/41.7 & 54.8/47.3/39.2  & 64.7/56.8/48.9  & 53.6/48.6/39.5  & 59.6/50.5/42.3 
\\

\hline
DeepMix & Hybrid &  \textbf{78.3}/\textbf{69.7}/\textbf{61.4} & \textbf{73.5/74.2/52.7} & \textbf{68.7/58.9/51.2} & \textbf{64.5/68.4/49.8} & \textbf{71.2/61.8/53.8}
\\
\end{tabular}
\caption{\small 
Class-wise 3D IoU@0.25 comparison of \name and state-of-the-art 3D object detection models for \hd{Scenario 3}.}
\label{tab:3DIoU-mp3}
\end{table*}

\begin{table}[]
\centering
\small 
\begin{tabular}{c|c|cccc|c}

\multirow{2}{3.5em}{Methods} & \multirow{2}{2.1em}{Input} & \multicolumn{4}{c|}{3D IoU@0.25: speed@0.5 m/s} & \multirow{2}{2.5em}{Average} \\ \cline{3-6}   
 & &\centering Chair  &\centering Bottle &\centering Box  &\centering Bag  &  \\    \hline 

MLCVNet~\cite{DBLP:conf/cvpr/XieLWWZXW20} & PC  &  54.2 & 56.1  & 47.1 & 37.5 & 48.7                                 


\\

AM3D~\cite{DBLP:conf/iccv/MaWLZOF19}    & RGB &  61.7 & 53.6  & 51.7  & 57.2  & 58.2
\\
D$^4$LCN~\cite{DBLP:conf/cvpr/DingHYWSLL20} & RGB &   59.7 & 51.2  & \textbf{64.1}  & 50.6  & 56.4 
\\

\hline
DeepMix & Hybrid &  \textbf{63.1} & \textbf{58.1} & 59.4 & \textbf{57.9} & \textbf{60.5}
\\
\end{tabular}
\caption{\small 
3D IoU@0.25 comparison of \name and state-of-the-art 3D object detection models for \hd{Scenario 1}.}
\label{tab:3DIoU-mp1}
\end{table}

\begin{table}[]
\centering
\small 
\begin{tabular}{c|c|cccc|c}

\multirow{2}{3.5em}{Methods} & \multirow{2}{2.1em}{Input} & \multicolumn{4}{c|}{3D IoU@0.25: speed@0.5 m/s} & \multirow{2}{2.5em}{Average} \\ \cline{3-6}   
 & &\centering Chair  &\centering Bottle &\centering Box  &\centering Bag  &  \\    \hline 


MLCVNet~\cite{DBLP:conf/cvpr/XieLWWZXW20} & PC  &  51.8 & 58.6  & 42.5 & 36.9 & 47.5                                
\\



AM3D~\cite{DBLP:conf/iccv/MaWLZOF19}    & RGB &  \textbf{67.2} & 59.4  & 60.4  & 55.4  & 60.6
\\
D$^4$LCN~\cite{DBLP:conf/cvpr/DingHYWSLL20} & RGB &   65.2 & 57.4  & \textbf{68.2}  & 52.4  & 60.8 
\\

\hline
DeepMix & Hybrid &  66.5 & \textbf{77.5} & 62.2 & \textbf{58.3} & \textbf{62.1}
\\
\end{tabular}
\caption{\small 
3D IoU@0.25 comparison of \name and state-of-the-art 3D object detection models for \hd{Scenario 2}.}
\label{tab:3DIoU-mp2}
\end{table}

Since AR/MR headsets are usually used in dynamic environments, we design three mobility patterns, as shown in Figure~\ref{fig:patterns}, to further evaluate the accuracy of \name.

\BULLET \hd{\hd{Scenario 1}}: The user moves along a 2-meter line, perpendicular to the line that connects its center and the object.

\BULLET \hd{Scenario 2}: The user moves along a 2-meter line, away, or toward the object (2 meters between line center and object).

\BULLET \hd{Scenario 3}: The user moves around the object in a circle with a diameter of 2 meters. 

In the mobile scenarios, we cannot replay the collected RGB-D images in our dataset that were captured at fixed locations.
Thus, we conduct controlled, live experiments with three moving speeds, 0.5, 1, and 2 m/s.
The user always looks at the object when moving with different patterns.

We first examine the 3D IoU results for \hd{Scenario 3} that are presented in Table~\ref{tab:3DIoU-mp3}.
For each setup with different movement patterns, moving speeds, 3D object detection methods, we run the experiments 20 times to measure detection accuracy.
Due to the large parameter space (\ie 8 methods, 3 patterns, 3 speeds, and 20 times for each setup), we select 4 out 7 classes, chair, bottle, box, and bag.
We only report the 3D IoU for the threshold of 0.25, since \name does not achieve the most accurate detection for only the chair and bottle categories with that threshold for the static scenario  (Table~\ref{tab:3DIoU-static}).

Table~\ref{tab:3DIoU-mp3} demonstrates that \name outperforms all existing 3D object detection models, for all four object classes, and under all three moving speeds.
The reason is that when the user moves around the object, \name can estimate and fine-tune the 3D bounding box from different viewing angles, significantly boosting the detection accuracy (\S\ref{sec:refine}).
Another key observation from this table is that the end-to-end latency drastically affects the detection accuracy in mobile scenarios.
While the best performing point-cloud-based model MLCVNet~\cite{DBLP:conf/cvpr/XieLWWZXW20} achieves comparable detection accuracy as the best image-based model D$^4$LCN~\cite{DBLP:conf/cvpr/DingHYWSLL20} for the static scenario (63.1\% vs. 63.9\% in Table~\ref{tab:3DIoU-static}), the detection accuracy of MLCVNet is much worse than D$^4$LCN for this mobile scenario (\eg 49.8\% vs. 59.6 when the moving speed is 0.5 m/s).
The worse performance of MLCVNet is mainly caused by its high end-to-end latency than D$^4$LCN (311 vs. 154 ms in Figure~\ref{fig:e2e}), which leads to the mismatch between objects and their bounding boxes due to accumulated tracking errors.

Table~\ref{tab:3DIoU-mp3} also shows that fast moving speeds reduce detection accuracy.
For example, the accuracy is 71.2\%, 61.8\%, and 53.8\% for the 0.5, 1, and 2 m/s moving speeds, respectively.
After analyzing the captured traces of different speeds, we find that high moving speeds can result in a larger and faster change of the object in the user's viewport than low moving speeds, which reduces the detection accuracy.

We next examine 3D IoU results for \hd{Scenario 1} in Table~\ref{tab:3DIoU-mp1} and \hd{Scenario 2} in Table~\ref{tab:3DIoU-mp2}, respectively.
In these tables, we show results for only 0.5 m/s moving speed.
For the other speeds, \name still outperforms existing models.
Moreover, we present results for only MLCVNet~\cite{DBLP:conf/cvpr/XieLWWZXW20} (the best performing point-cloud-based method), AM3D~\cite{DBLP:conf/iccv/MaWLZOF19}, and D$^4$LCN~\cite{DBLP:conf/cvpr/DingHYWSLL20}.
D$^4$LCN performs better than \name for boxes.
One of the possible reasons is that some boxes in our collected dataset have uneven surfaces (\eg big holes, Figure~\ref{fig:workflow}) that affect the quality of captured depth images, which the lightweight  scheme~\cite{grunnet2018depth} in \name cannot fix.
This issue could potentially be addressed by leveraging computer graphics algorithms (\eg surface reconstruction~\cite{izadi2011kinectfusion}).
The performance of AM3D is slightly better than \name for the chair category for only \hd{Scenario 2} (67.2\% vs. 66.5\%).
For \hd{Scenario 1}, \name still outperforms AM3D (63.1\% vs. 61.7\%).
The comparison between Table~\ref{tab:3DIoU-mp1} and Table~\ref{tab:3DIoU-mp2} reveals that \hd{Scenario 2} leads to better performance than \hd{Scenario 1} (\eg 62.1\% vs. 60.5\% for \name).
The reason is that, similar to the case of different moving speeds, \hd{Scenario 1} may lead to a larger and faster change of the object in the viewport than \hd{Scenario 2}, affecting detection accuracy.

By comparing Table~\ref{tab:3DIoU-mp3} with Tables~\ref{tab:3DIoU-mp1} and~\ref{tab:3DIoU-mp2}, we find that the 3D object detection accuracy is better for \hd{Scenario 3} than the other two scenarios (\eg 71.2\% vs. 60.5\% and 62.1\% for \name at 0.5 m/s moving speed.).
One possible reason is that moving around the object leads to more opportunities to view it from different directions than the other two mobility patterns, which refine the detection accuracy of \name (\S\ref{sec:refine}).
Note that existing 3D object detection models cannot benefit from the movement of users.
For example, the detection accuracy is 59.6\%, 60.8\%, and 56.4\% for D$^4$LCN~\cite{DBLP:conf/cvpr/DingHYWSLL20} for scenarios 3, 2, and 1, respectively.
On the other hand, their accuracy may drop when users are moving around, due to their high end-to-end latency of 3D object detection.
%

\mysubsection{Impact of Dynamic Network Conditions}
\label{sec:network}

In addition to the high-throughput WiFi network, we also evaluate the performance of \name on our USRP-based LTE testbed with fluctuating network bandwidth (8.4--37.1 Mbps vs. 260 Mbps for WiFi).
The latency of LTE is also higher than that of WiFi (14 vs. 1 ms).
The dynamic network conditions affect the end-to-end latency of \name (\ie longer data transmission time and higher network latency).
We compare the performance of \name with three models Mono3D~\cite{DBLP:conf/cvpr/ChenKZMFU16}, Trans3D~\cite{DBLP:conf/iccv/TangL19}, and COG~\cite{ren2016three}, which have the lowest end-to-end latency in their own categories (Figure~\ref{fig:e2e}).

We plot the end-to-end latency on the LTE network in Figure~\ref{fig:LTE}.
In this figure, the latency of data transmission on the LTE network is more visible, compared to the results in Figure~\ref{fig:e2e} for the WiFi network.
The end-to-end latency of \name increases from 34 ms to 47--62 ms (51 on average) and is dominated by data transmission (on average 22 ms, 43.1\%), whereas the latency of Mono3D~\cite{DBLP:conf/cvpr/ChenKZMFU16}, which has the lowest latency among existing methods, increases from 91 ms to 105--118 ms, higher than the 100 ms latency requirement of interactive AR/MR~\cite{Chen:2018:MEM,Liu:2019:EAR}.
The data transmission time increases from 5 ms to 18--33 ms for RGB images, and from 8 ms to 24--39 ms for RGB-D images.

The headset is static during the experiments, and thus the dynamic network conditions have a limited impact on detection accuracy.
On this USRP-based LTE network, moving the headset attached with an LTE dongle makes network connection unstable.
Hence, we do not conduct mobile experiments.

\mysubsection{User Study}
\label{sec:user-study}

Besides the above evaluations on datasets and controlled experiments, we assess the performance of \name through an IRB-approved user study.
Our goal is to understand how the smoothness and accuracy of 3D object detection affect user experience.
We define smoothness as the update frequency of 3D bounding boxes in dynamic environments, which reflects the end-to-end latency of object detection.

We conduct the user study with 33 diverse participants with age 18 to 45.
Among them, 7 are female, 21 are familiar with AR/MR, and 3 used mobile headsets before.
We ask each participant to experience three 3D object detection schemes, \name, MLCVNet~\cite{DBLP:conf/cvpr/XieLWWZXW20} (point-cloud-based), and D$^4$LCN~\cite{DBLP:conf/cvpr/DingHYWSLL20} (image-based).
The last two have better accuracy than other existing solutions (Table~\ref{tab:3DIoU-static}, Table~\ref{tab:3DIoU-mp1}, and Table~\ref{tab:3DIoU-mp2}).
We randomly order the three schemes, and thus participants do not know which one is \name.
We ask the participants to compare the smoothness and accuracy of the three solutions by providing their mean opinion scores (MOS), from 1 to 5 (1: bad, 2: poor, 3: fair, 4: good, 5: excellent). 
Participants experience the detection of a chair and a bottle by following the three mobility patterns in Figure~\ref{fig:patterns}.

We observe the following from the results of our user study in Figure~\ref{fig:user-study}.
First, \name leads to the best user experience in terms of both smoothness and accuracy, thanks to its lightweight and accurate 3D object detection.
For the average score, the smoothness of \name is 44.3\% (10.4\%), 59.4\% (16.6\%), and 56.9\% (11.7\%) higher than that of MLCVNet~\cite{DBLP:conf/cvpr/XieLWWZXW20} (D$^4$LCN~\cite{DBLP:conf/cvpr/DingHYWSLL20}) for the three mobility patterns; whereas the accuracy of \name is 48.2\% (12.5\%), 65.4\% (5.1\%), and 52.3\% (9.6\%) higher than that of MLCVNet~\cite{DBLP:conf/cvpr/XieLWWZXW20} (D$^4$LCN~\cite{DBLP:conf/cvpr/DingHYWSLL20}) for the three patterns.
Second, the point-cloud-based model MLCVNet~\cite{DBLP:conf/cvpr/XieLWWZXW20} has the worse performance among the three, mainly caused by its high end-to-end latency (due to the huge amount of 3D data to process).
This demonstrates the importance of end-to-end latency for mobile AR/MR.
Third, for both smoothness and accuracy, the average score of \name is still slightly lower than 4 (good experience), which shows that there is room to improve for achieving immersive experience on mobile headsets.

\mysubsection{Effectiveness of Bounding Box Caching}
\label{sec:box-caching}

We conduct controlled experiments to evaluate the 3D bounding box caching and reusing scheme  (\S\ref{sec:caching}). 
We mount two headsets on a gimbal that can rotate at a fixed angular velocity.
Both headsets are equipped with \name, but only one of them has caching enabled.
We randomly place 3 objects around the gimbal for testing and rotate both headsets $720^{\circ}$ at the same time with the same speed. 
The result shows that the time to render the bounding box of a cached item is only 2.6--3.2 ms.
Without caching, \name needs to execute the entire workflow, which takes at least 34 ms.
Moreover, the number of offloaded frames decreases from 113 to 12 when caching is enabled and almost does not change without caching. 
Thus, our bounding box caching optimization can drastically improve user experience, reduce the amount of offloaded data, and decrease computation overhead on both the edge and the headset.

\mysubsection{Temperature, Power, and Computation Resources}
\label{sec:resource}

To demonstrate its lightweight feature, we finally compare the on-device surface temperature, battery power level, and computation resource utilization of \name with MLCVNet~\cite{DBLP:conf/cvpr/XieLWWZXW20} (point-cloud-based), and D$^4$LCN~\cite{DBLP:conf/cvpr/DingHYWSLL20} (image-based).
Existing schemes in the same category have almost the same performance, because their heavy-lifting jobs are all offloaded to the edge.
We generate point clouds for MLCVNet on mobile headsets to demonstrate its overhead.
Otherwise, the performance of MLCVNet is close to that of D$^4$LCN.
We record the surface temperature of the headset after 40-minute use for each scheme and plot the results in Figure~\ref{fig:temp}. 
Since \name needs process depth data in real-time on the headset for estimating the 3D bounding box, the surface temperature is slightly higher than that of D$^4$LCN. 
Due to the high cost of generating point clouds, in the end, the surface temperature for MLCVNet is about 
3.4$^{\circ}C$ higher than that of \name and D$^4$LCN. %
After using the headset for 40 minutes, the average battery power level of \name is 8.2 W, which is 0.3 W higher than D$^4$LCN and 1.8 W lower than MLCVNet.
The average CPU (memory) usage of \name is only 24.3\% (11.3\%), which is 2.3\% (0.4\%) higher than D$^4$LCN and 7.6\% (6.4\%) lower than MLCVNet.
%
%
%
There is no significant difference in GPU usage among these methods.
Unfortunately, we do not find a method to measure the HPU utilization of Microsoft HoloLens 2.
\mysection{Discussion and Future Work} 
\label{sec:discussion}

\noindent {\bf Image-based 3D Object Detection.}
As shown in Table~\ref{tab:3DIoU-static}, image-based 3D object detection such as D$^4$LCN~\cite{DBLP:conf/cvpr/DingHYWSLL20} achieves high accuracy for the static scenario (\eg only 3.5\% lower than \name for 3D IoU@0.25).
Its poor performance for the dynamic scenario (11.6\% lower than \name in Table~\ref{tab:3DIoU-mp3} for speed@0.5 m/s) is mainly caused by the high end-to-end latency.
However, it can potentially handle more use cases than \name, as we will discuss next.
We plan to optimize the runtime inference performance of image-based 3D object detection to make it practical.

\noindent {\bf Limitations.}
As the first-of-its-kind accurate 3D object detection that is suitable for mobile headsets, \name has a few limitations of its current design.
For example, it can detect mainly objects that are placed on a plain surface, and it is challenging to detect, for instance, a TV that is hung on a wall.
%
%
However, we argue that the target scenarios of \name (\eg objects on a plane) 
are the common use cases for indoor mobile AR/MR.
Another issue is that the range of depth cameras is typically limited (\eg from 0.5 to 5.5 m), which our caching scheme can help only to some extent.
Hence, \name cannot detect objects that are far away from users.
We are extending \name to address these limitations.
One possible solution is to design a hybrid scheme that dynamically switches between \name and image-based 3D object detection, given that plane detection is a solved problem~\cite{arkit,fischler1981random,arcore} and the range of RGB cameras is longer than that of depth ones.

\noindent {\bf Supporting Interactive AR/MR.} The lightweight, accurate 3D object detection offered by \name lays the foundation for enabling real-time, interactive AR/MR on mobile headsets.
%
We plan to build various immersive applications by leveraging this key capability of \name. 
\mysection{Related Work}
\label{sec:related}

In this section, we review the related work on 3D object detection and mobile AR/MR. 


%

\vspace{0.05in} 

\noindent \textbf{3D Object Detection.}
{\em Point-cloud-based 3D Object Detection: } 
With the development of deep learning models on point clouds~\cite{DBLP:conf/cvpr/QiSMG17}, many 3D object detection models have emerged~\cite{engelcke2017vote3deep,lang2019pointpillars,qi2019deep,DBLP:conf/cvpr/QiSMG17,DBLP:conf/cvpr/XieLWWZXW20,DBLP:conf/cvpr/ZhouT18}. 
For example, approaches such as 
VoteNet~\cite{qi2019deep}, COG~\cite{ren2016three}, and MLCVNet~\cite{DBLP:conf/cvpr/XieLWWZXW20} can directly take raw point cloud as input. 
Thanks to the available depth information and the underlying DNN networks, those methods can achieve high detection accuracy. 
\bo{
}
{\em Image-based 3D Object Detection:} 
Instead of processing point clouds, some existing schemes~\cite{DBLP:conf/cvpr/ChenKZMFU16,chen20153d,DBLP:conf/cvpr/DingHYWSLL20,DBLP:conf/iccv/MaWLZOF19} utilize 2D detectors to achieve 3D object detection.
%
%
%
%
%
For example, D$^4$LCN~\cite{DBLP:conf/cvpr/DingHYWSLL20} estimates the depth information from monocular images and fuses RGB and depth using improved 2D convolutions to generate 3D bounding boxes.
%
%
{\em 3D Object Detection with RGB-D Input: }
This category utilizes both RGB images and depth data for 3D object detection~\cite{lahoud20172d,qi2017frustum,song2016deep,DBLP:conf/iccv/TangL19}.
For example, F-PointNet~\cite{qi2017frustum} narrows down the 3D space by leveraging 2D object detection and further performs segmentation on selected 3D frustums with PointNet~\cite{DBLP:conf/cvpr/QiSMG17} to help estimate the 3D bounding box.
%
%
Although these approaches can reduce the amount of to-be-processed 3D data, their accuracy is usually not as good as point-cloud-based schemes.
%
%
%
Different from the above work, \name benefits from 2D object detection models that have low computation latency. 
By utilizing real-time depth information from sensors, it can achieve {\em high 3D object detection accuracy with low end-to-end latency}.

\vspace{0.05in}
\noindent {\bf Mobile AR/MR.}
There is a rich literature on building mobile AR/MR systems~\cite{apicharttrisorn2019frugal,Chen:2018:MEM,choi2019lpgl,kang2018my,li2019holodoc,park2018mixed,yi2020heimdall}.
For example, 
%
%
HomeMeld~\cite{kang2018my} enables the telepresence between remote living areas through robot agents as avatars, by finding an equivalent functional place in rooms and predicting real-time paths to prevent lagging caused by the robot’s slow movement.
%
%
LpGL~\cite{choi2019lpgl} is a device-independent graphics library that reduces energy consumption for mobile headset applications, which dynamically selects frame rate and object shape complexity and leverages user movements to extend the battery life.
Heimdall~\cite{yi2020heimdall} coordinates concurrent GPU usage for multi-tasking in mobile AR applications, by splitting the DNNs into small units and executing them between rendering frames.
Different from the above work, \name offers a  real-time, accurate 3D object detection framework, which is {\em missing in existing mobile AR/MR systems}, to support better interaction between and seamless integration of the digital and 3D physical worlds and provide a truly immersive user experience for headset-based applications.


\mysection{Conclusion}

In this paper, we present the design, implementation, and evaluation of \name, a mobility-aware, lightweight, and accurate 3D object detection system for improving the quality of user experience of AR/MR applications running on mobile headsets.
Instead of directly leveraging/accelerating existing 3D object detection models that are computation-intensive, \name benefits from mature 2D object detection algorithms to derive a bounding box for the object of interest.
It then utilizes this 2D bounding box to extract depth data from depth images captured by the headset and estimates the 3D bounding box by effectively exploring 3D geometry and data processing.
By doing this, \name not only reduces the end-to-end latency of AR/MR applications but also drastically increases the detection accuracy in dynamic environments, by fully exploiting the mobility of headsets.
We implement \name on a commodity mobile headset and compare its performance with several state-of-the-art 3D object detection models.
Our extensive experiments, including a user study, demonstrate the efficacy of \name in terms of both end-to-end latency and detection accuracy.


\small
\bibliographystyle{abbrv}
\bibliography{headset}

\end{document}